\documentclass{article}
\usepackage[numbers,sort&compress]{natbib}
\usepackage{neurips_2024}
\usepackage{amsmath, amsfonts, amssymb}
\usepackage{booktabs}
\usepackage{graphicx}
\usepackage{wrapfig}
\usepackage{tikz}
\usetikzlibrary{shapes.geometric, shapes.misc, arrows.meta, positioning, calc, backgrounds, fit}
\usepackage{hyphenat}
\usepackage[utf8]{inputenc}
\usepackage{kotex}
\usepackage{microtype}
\usepackage{subcaption}
\usepackage[table]{xcolor}
\usepackage{makecell}

\title{Motif 3: Technical Report}
\author{\textbf{Motif Technologies}}

\begin{document}
\maketitle

\begin{abstract}

We introduce Motif 3, a decoder-only Mixture-of-Experts language model
with 314 billion total parameters and 13.2 billion parameters activated
for each token. Each sparse MoE layer contains 384 routed experts, while
only eight are selected foreach token. This fine-grained sparsity
provides substantial expert capacity while limiting the expert
computation required for each token. Motif 3 is built around Grouped
Differential Latent Attention (GDLA), which integrates grouped differential attention with
the compressed key-value representation of Multi-head Latent Attention.
The architecture further incorporates modified manifold-constrained
hyper-connections, Expert-Specific PolyNorm activations, and multi-token
prediction to improve optimization stability, expert specialization,
and inference efficiency. We pretrain Motif 3 on approximately 12.5
trillion tokens spanning web documents, STEM, code, mathematics,
multilingual content, and domain-specialized corpora. A collection of
expert-balancing and numerical-stabilization techniques supports stable
training at scale, while selective MXFP8 computation and communication,
memory-efficient fused kernels, and window-aware context parallelism
enable training with context lengths up to 256K tokens. Our post-training
pipeline combines general supervised fine-tuning, six specialist
teachers trained with reinforcement learning, a software-engineering
teacher trained with supervised fine-tuning, and Multi-teacher On-Policy
Distillation. The resulting unified model consolidates complementary
capabilities in reasoning, coding, tool use, professional work,
long-context understanding, calibrated abstention, and instruction
following. Across a broad evaluation suite, Motif 3 demonstrates
competitive performance against leading open-weight models, including
strong results on long-horizon agentic tasks, mathematical reasoning,
scientific knowledge, and hallucination-sensitive evaluation.

\end{abstract}

\section{Introduction}

Large language models (LLMs) have rapidly evolved from general-purpose language systems into increasingly capable problem solvers, demonstrating substantial progress in reasoning, coding, tool use, and long-horizon agentic tasks~\cite{qwen3, kimi_k2_5, glm5, deepseek_v4}. In parallel, recent open-weight models have continued to narrow the capability gap with leading proprietary systems, while Mixture-of-Experts (MoE) architectures have enabled the scaling of model capacity without proportionally increasing the computational cost incurred for each token. These advances highlight that further progress in foundation models depends not only on scaling data and computation, but also on developing more expressive and efficient architectures that can translate large-scale training into strong and broadly generalizable intelligence.

In this work, we introduce \textbf{Motif 3}, a Mixture-of-Experts language model with 314 billion total parameters and 13.2 billion parameters activated for each token. A defining feature of Motif 3 is its fine-grained sparse expert design: each MoE layer contains 384 routed experts, of which only eight are selected for each token. This design provides a large pool of expert capacity while limiting the expert computation incurred by each token. At the core of Motif 3 is \textbf{Grouped Differential Latent Attention} (GDLA), a novel attention architecture that combines our grouped formulation of differential attention~\cite{motif2025grouped}, building on Differential Transformer~\cite{ye2024differential}, with Multi-head Latent Attention (MLA)~\cite{deepseekv2}. GDLA is designed to retain the expressive attention dynamics enabled by differential attention while substantially reducing the key-value cache requirements through latent representations. Motif 3 further incorporates manifold-constrained hyper-connections (mHC)~\cite{xie2025mhc} and the gated-attention mechanism adopted in Qwen3-Next~\cite{qiu2025gated,qwen3next}, forming an architecture designed for efficient inference and stable optimization at scale.

We pretrain Motif 3 on approximately 12.5 trillion tokens drawn from a diverse mixture of high-quality web documents, STEM content, source code, mathematical data, synthetic question-answer pairs, and domain-specialized corpora, with additional emphasis on Korean and multilingual data, reasoning-intensive examples, and legal and financial domains. To train this large expert pool reliably, we employ a layered expert-balancing strategy that combines sigmoid-based routing, auxiliary-loss-free expert biases, sequence-wise load-balancing objectives, decaying router noise, and layer-adaptive auxiliary-loss coefficients. Together, these mechanisms encourage broad expert exploration during the early stages of training while preventing persistent expert overload, starvation, and specialization collapse as training progresses. We additionally anneal the mHC post-mapping multiplier from its original value of two to one to limit the accumulation of residual-stream activation outliers.

Starting from the pretrained model, we employ a staged post-training
pipeline comprising general Supervised Fine-Tuning (SFT), specialized
teacher training, and Multi-teacher On-Policy Distillation
(MOPD)~\cite{xiao2026mimo,yang2026nemotron3ultra}. We first use a
preliminary SFT model to identify capability-specific failure modes and
generate targeted supervision, particularly for failure-prone decisions
in agentic trajectories. We then restart from the pretrained checkpoint
and train the general SFT model on the consolidated corpus, combining
these targeted examples with complete interaction trajectories and
broader instruction, reasoning, coding, and knowledge data. From this
general SFT checkpoint, we train six specialist teachers with
domain-specific GRPO across 13 verifier domains and construct a seventh,
software-engineering teacher through SFT. MOPD then consolidates the
complementary capabilities of all seven teachers into the general
student, producing a single model with broad coverage and targeted
expertise.

The remainder of this report is organized as follows. We first describe
the Motif 3 architecture, including GDLA, the MoE design, mHC, and MTP.
We then present the system optimizations used for low-precision,
distributed, and long-context training, followed by the pretraining data,
tokenizer, training configuration, and MoE stabilization methods. Next, we
describe general SFT, specialized teacher training, and MOPD. Finally,
we evaluate Motif 3 across agentic, reasoning, knowledge, instruction-
following, and long-context benchmarks.
\section{Architecture}

\subsection{Overview}
\label{sec:architecture_overview}

\begin{figure}[t!]
    \centering
    \input{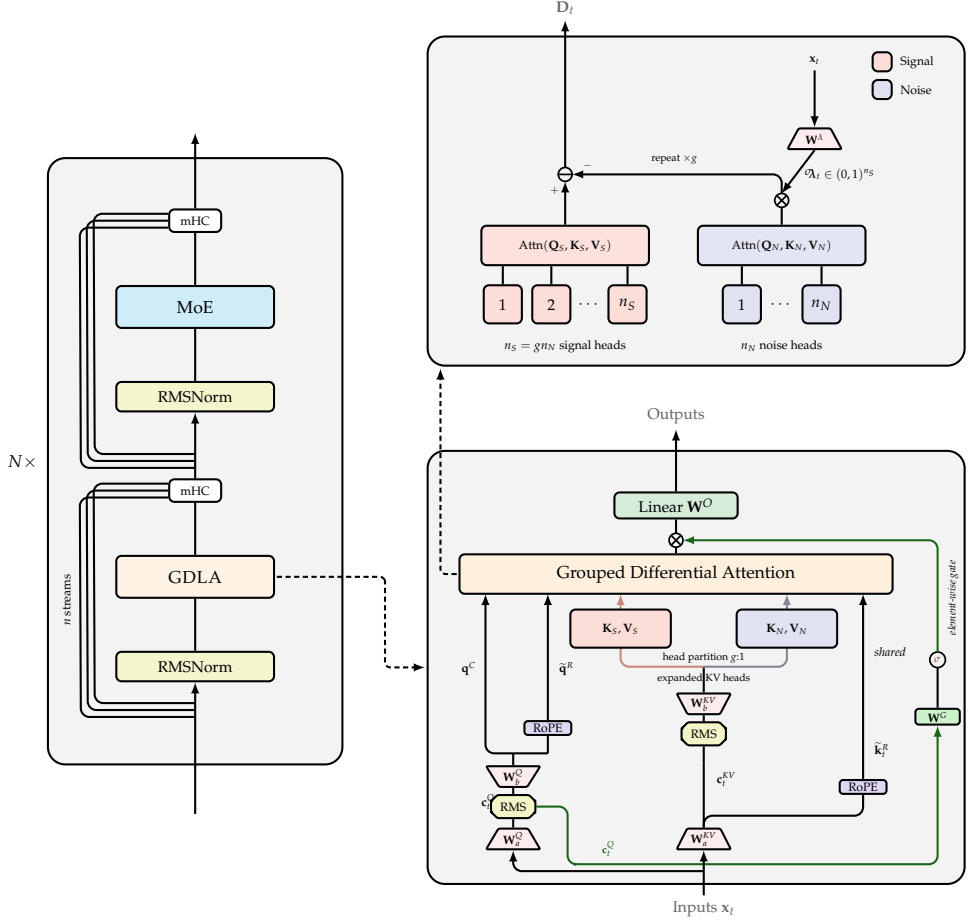}
    \caption{\small
        Illustration of the GDLA (\textbf{G}rouped \textbf{D}ifferential
        \textbf{L}atent \textbf{A}ttention) architecture of Motif~3.
        \textbf{Left}: each Transformer block pairs a GDLA token-mixing
        layer with a sparse MoE channel-mixing layer and uses mHC to mix
        $n$ parallel residual streams.
        \textbf{Bottom right}: the query and KV paths use low-rank
        projections. The complete KV latent $\mathbf{c}^{KV}$ is
        RMS-normalized and expanded once by $\mathbf{W}^{KV}_{b}$ into
        16 KV heads shared by the signal and noise query paths. The
        decoupled rotary key is also shared by all KV heads. An
        element-wise gate computed from the normalized query latent
        $\mathbf{c}^{Q}$ modulates the differential output before
        $\mathbf{W}^{O}$.
        \textbf{Top right}: $n_S=g n_N$ signal heads are paired with
        repeated noise-head outputs. A token-dependent coefficient
        $\boldsymbol{\lambda}_t=\sigma\!\left(\mathbf{x}_t\mathbf{W}^{\lambda}\right)
        \in(0,1)^{n_S}$ is predicted for every signal head and scales the
        noise term before subtraction.
    }
    \label{fig:gdla_architecture}
\end{figure}

Motif 3 is a decoder-only autoregressive Mixture-of-Experts (MoE)
language model with 314 billion total parameters and approximately
13.2 billion parameters activated for each token.
A defining feature of the model is its fine-grained sparsity: each
sparse MoE layer contains 384 routed experts and selects eight for each
token. This design expands the available expert capacity without a
proportional increase in per-token expert computation.
As illustrated in Figure~\ref{fig:gdla_architecture}, the model is
composed of repeated Transformer blocks, each containing a
Grouped Differential Latent Attention (GDLA) layer followed by a
sparse MoE feed-forward layer.
We use a hybrid GDLA schedule that repeats one full causal-attention
layer followed by three sliding-window attention layers. Both layer
types use the same GDLA formulation.
We apply RMSNorm~\cite{zhang2019rmsnorm} before each sublayer and use
modified manifold-constrained hyper-connections
(mHC)~\cite{xie2025mhc} in place of conventional residual additions.

GDLA combines Grouped Differential Attention
(GDA)~\cite{motif2025grouped} with Multi-head Latent Attention
(MLA)~\cite{deepseekv2,deepseekv3} and applies a query-dependent output
gate~\cite{qiu2025gated,qwen3next}. The sparse MoE feed-forward layers
use Expert-Specific PolyNorm activations. We additionally adopt multi-token
prediction (MTP) following DeepSeek-V3~\cite{deepseekv3}. The following
sections describe these components in detail.

\subsection{Grouped Differential Latent Attention}
\label{sec:gdla}

Standard self-attention can allocate substantial probability mass to
irrelevant or redundant context. Differential Attention improves
attention selectivity by subtracting one attention distribution from
another, canceling patterns shared by the signal and noise paths and
concentrating attention on more relevant context
~\cite{ye2024differential}. Its symmetric formulation, however,
allocates equal head capacity to signal modeling and noise estimation.
GDA retains the noise-suppression mechanism while assigning more heads
to the signal path and sharing a smaller set of noise heads through
controlled repetition. This increases signal-modeling capacity with
limited additional computation~\cite{motif2025grouped}.

\begin{figure}[t]
    \centering
    \includegraphics[width=0.78\textwidth]{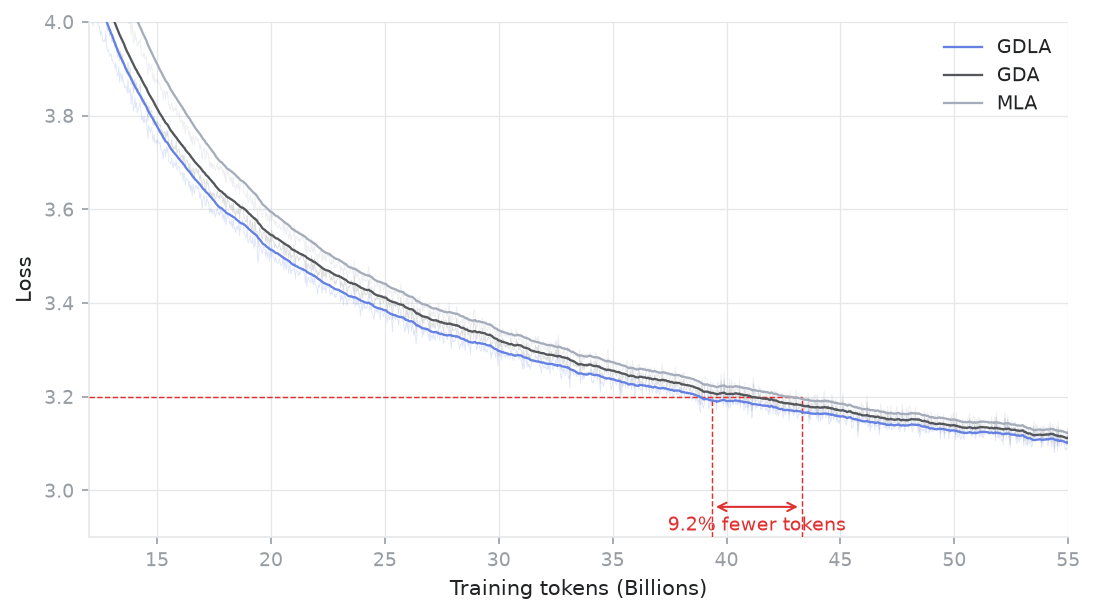}
    \caption{
        Attention training-loss comparison.
        GDLA achieves lower loss than GDA and MLA and reaches a loss of
        $3.2$ with $9.2\%$ fewer training tokens than MLA.
    }
    \label{fig:attn_loss_comparison}
\end{figure}

MLA addresses a complementary efficiency bottleneck by compressing the
KV state into a low-rank latent representation, substantially reducing
the KV-cache requirements of autoregressive inference
~\cite{deepseekv2,deepseekv3}. GDLA combines these advantages: it retains
differential noise suppression, uses the asymmetric GDA allocation for
efficient signal modeling, and preserves MLA's compact inference state.
It applies a single normalization and up-projection to the compressed KV
latent and separates the signal and noise paths only along the expanded
head dimension.
In a controlled training comparison, GDLA maintains a lower loss than
both GDA and MLA and reaches a loss of $3.2$ with $9.2\%$ fewer training
tokens than MLA, as shown in Figure~\ref{fig:attn_loss_comparison}.
All three diagnostic comparisons reported in
Figures~\ref{fig:attn_loss_comparison}
and~\ref{fig:moe_component_analysis} were measured in controlled
experiments using models with approximately 10 billion parameters.

\subsubsection{Latent Query and Key-Value Representations}
\label{sec:gdla_latent}

Let $\mathbf{x}_t \in \mathbb{R}^{d}$ denote the hidden representation
at token position $t$.
Following MLA, the query is constructed through a low-rank
down-projection and up-projection:

\begin{align}
    \mathbf{c}^{Q}_t
    &=
    \operatorname{RMSNorm}
    \left(
        \mathbf{x}_t \mathbf{W}^{Q}_{a}
    \right),
    \\
    \left[
        \mathbf{q}^{C}_t;
        \mathbf{q}^{R}_t
    \right]
    &=
    \mathbf{c}^{Q}_t \mathbf{W}^{Q}_{b},
\end{align}

where $\mathbf{q}^{C}_t$ and $\mathbf{q}^{R}_t$ denote the content and
rotary query components, respectively.

The KV down-projection jointly produces a compressed KV representation
and a decoupled rotary key:

\begin{equation}
    \left[
        \mathbf{c}^{KV}_t;
        \mathbf{k}^{R}_t
    \right]
    =
    \mathbf{x}_t \mathbf{W}^{KV}_{a}.
\end{equation}

The full KV latent is normalized and expanded through a single
projection:

\begin{align}
    \overline{\mathbf{c}}^{KV}_t
    &=
    \operatorname{RMSNorm}
    \left(
        \mathbf{c}^{KV}_t
    \right),
    \\
    \left[
        \mathbf{K}^{C}_t;
        \mathbf{V}_t
    \right]
    &=
    \operatorname{Reshape}_{\mathrm{heads}}
    \left(
        \overline{\mathbf{c}}^{KV}_t
        \mathbf{W}^{KV}_{b}
    \right).
\end{align}

The 16 expanded content-key and value heads are shared by the signal
and noise query paths. Thus the two paths differ in their query heads
while using a common compressed KV representation. This avoids
maintaining separate KV states while retaining a compact state for
inference.

RoPE~\cite{su2021roformer} is applied only to the rotary query and key
components:

\begin{align}
    \widetilde{\mathbf{q}}^{R}_t
    &=
    \operatorname{RoPE}_{t}
    \left(
        \mathbf{q}^{R}_t
    \right),
    \\
    \widetilde{\mathbf{k}}^{R}_t
    &=
    \operatorname{RoPE}_{t}
    \left(
        \mathbf{k}^{R}_t
    \right).
\end{align}

The same rotary key is broadcast across all KV heads:

\begin{equation}
    \mathbf{K}_{t}
    =
    \left[
        \mathbf{K}^{C}_{t};
        \operatorname{Repeat}
        \left(
            \widetilde{\mathbf{k}}^{R}_t
        \right)
    \right].
\end{equation}

The two paths consequently operate under a common positional reference.

\subsubsection{Grouped Differential Attention}
\label{sec:gdla_grouped_differential}

Let $n_S$ and $n_N$ denote the numbers of signal and noise query heads.
For a total of $n_H$ query heads, we use

\begin{align}
    n_N
    &=
    \frac{n_H}{g+1},
    &
    n_S
    &=
    g n_N.
\end{align}

where $g$ is the grouped ratio.
The signal and noise query paths share the same KV heads. When fewer KV
heads than query heads are used, the KV heads are associated with
multiple query heads following grouped-query
attention~\cite{ainslie2023gqa}.
This arrangement assigns most of the attention capacity to signal
modeling while limiting the additional computation required for noise
estimation.

The signal and noise paths independently compute

\begin{align}
    \mathbf{H}_{S}
    &=
    \operatorname{Attn}
    \left(
        \mathbf{Q}_{S},
        \mathbf{K},
        \mathbf{V}
    \right),
    \\
    \mathbf{H}_{N}
    &=
    \operatorname{Attn}
    \left(
        \mathbf{Q}_{N},
        \mathbf{K},
        \mathbf{V}
    \right).
\end{align}

Because the signal path contains $g$ times as many heads, each noise
output is repeated across the corresponding signal-head group.
Unlike the static layer-wise coefficient used in the original
differential formulation, GDLA predicts an input-dependent coefficient
for every signal head:

\begin{equation}
    \boldsymbol{\lambda}_t
    =
    \sigma\!\left(\mathbf{x}_t \mathbf{W}^{\lambda}\right),
    \qquad
    \boldsymbol{\lambda}_t\in(0,1)^{n_S}.
\end{equation}

The sigmoid constrains each coefficient to $(0,1)$, so the differential
operation can suppress the noise path but never invert or amplify it.

The grouped differential output is then written as

\begin{equation}
    \mathbf{D}_t
    =
    \mathbf{H}_{S,t}
    -
    \boldsymbol{\lambda}_t
    \odot
    \operatorname{Repeat}_{g}
    \left(
        \mathbf{H}_{N,t}
    \right).
\end{equation}

Here, $\operatorname{Repeat}_{g}$ repeats each noise-head output $g$
times so that each signal head is paired with the noise head of its own
group.

\subsubsection{Output Gating}
\label{sec:gdla_output_gate}

Before the final output projection, the differential representation is
modulated by an element-wise sigmoid gate computed
from the normalized query latent:

\begin{equation}
    \mathbf{G}_t
    =
    \operatorname{Reshape}_{\mathrm{heads}}
    \left(
        \mathbf{c}^{Q}_t \mathbf{W}^{G}
    \right),
    \qquad
    \mathbf{G}_t\in\mathbb{R}^{n_S\times d_V},
\end{equation}

where
$\mathbf{W}^{G}\in\mathbb{R}^{r_Q\times(n_S d_V)}$ maps the
$r_Q$-dimensional query latent to one gate value for every element of
the $n_S$ signal-head outputs, and $d_V$ is the value-head dimension.
The final attention output at token $t$ is

\begin{equation}
    \operatorname{GDLA}(\mathbf{x}_t)
    =
    \operatorname{vec}
    \left(
        \sigma
        \left(
            \mathbf{G}_t
        \right)
        \odot
        \mathbf{D}_t
    \right)
    \mathbf{W}^{O}.
\end{equation}

Here,
$\mathbf{W}^{O}\in\mathbb{R}^{(n_S d_V)\times d}$ projects the
concatenated gated signal-head outputs back to the model dimension.

This gating mechanism allows the model to control individual attention
output channels as a function of the current hidden representation.

\subsection{Modified Manifold-Constrained Hyper-Connections}
\label{sec:modified_mhc}

Motif 3 replaces conventional residual additions with a modified form
of manifold-constrained hyper-connections (mHC)~\cite{xie2025mhc}.
mHC generalizes a single residual stream into multiple parallel streams
and learns token-dependent mappings that select, transform, and
redistribute information across them. This provides richer cross-layer
information flow than a fixed residual addition, while the manifold
constraints preserve a convex mixing structure and identity-like
propagation. Motif 3 retains these benefits but modifies the
post-mapping scale to prevent repeated residual amplification from
producing activation outliers at large depth. The resulting formulation
combines the expressive multi-stream connectivity of mHC with improved
training stability.

Let $\mathbf{X}_{\ell,t}\in\mathbb{R}^{n\times d}$ denote the $n$
parallel residual streams at token position $t$ and layer $\ell$.
We use an expansion rate of $n=4$, corresponding to four parallel
residual streams.
An mHC block first reduces these streams to the input of a sublayer and
then redistributes the sublayer output while mixing the original
streams:

\begin{equation}
    \mathbf{X}_{\ell+1,t}
    =
    \mathbf{H}_{\mathrm{res},\ell,t}
    \mathbf{X}_{\ell,t}
    +
    \mathbf{H}_{\mathrm{post},\ell,t}^{\top}
    F_{\ell}
    \left(
        \mathbf{H}_{\mathrm{pre},\ell,t}
        \mathbf{X}_{\ell,t};
        \mathbf{W}_{\ell}
    \right).
    \label{eq:mhc_block}
\end{equation}

Here, $\mathbf{H}_{\mathrm{pre}}\in\mathbb{R}^{1\times n}$
computes a weighted reduction across streams,
$\mathbf{H}_{\mathrm{post}}\in\mathbb{R}^{1\times n}$ scales the
sublayer output as it is broadcast back to the streams, and
$\mathbf{H}_{\mathrm{res}}\in\mathbb{R}^{n\times n}$ mixes the
residual streams directly.
The pre-mapping and post-mapping are nonnegative vectors.
The residual mapping is constrained to the Birkhoff polytope, so its
entries are nonnegative and every row and column sums to one.
This doubly stochastic constraint preserves a convex mixing structure
and restores the identity-mapping behavior that can be lost in
unconstrained hyper-connections.

All three mappings are generated dynamically at each token position.
We flatten the residual-stream and hidden dimensions, apply RMSNorm,
and use a single merged projection:

\begin{align}
    \mathbf{q}_{\ell,t}
    &=
    \operatorname{RMSNorm}
    \left(
        \operatorname{vec}(\mathbf{X}_{\ell,t})
    \right),
    \\
    \left[
        \mathbf{p}_{\mathrm{pre}};
        \mathbf{p}_{\mathrm{post}};
        \operatorname{vec}(\mathbf{P}_{\mathrm{res}})
    \right]
    &=
    \mathbf{W}_{\mathrm{mHC}}
    \mathbf{q}_{\ell,t}.
    \label{eq:mhc_dynamic_projection}
\end{align}

Learned scalar projection scales and biases transform these outputs.
The pre-mapping is computed with a sigmoid, while the residual mapping
is obtained through Sinkhorn-Knopp normalization:

\begin{align}
    \mathbf{H}_{\mathrm{pre}}
    &=
    \sigma
    \left(
        \alpha_{\mathrm{pre}}\mathbf{p}_{\mathrm{pre}}
        +\mathbf{b}_{\mathrm{pre}}
    \right),
    \\
    \mathbf{H}_{\mathrm{res}}
    &=
    \operatorname{Sinkhorn}
    \left(
        \exp
        \left(
            \alpha_{\mathrm{res}}\mathbf{P}_{\mathrm{res}}
            +\mathbf{B}_{\mathrm{res}}
        \right)
    \right).
    \label{eq:mhc_constrained_mappings}
\end{align}

The projection outputs, mapping logits, and Sinkhorn-Knopp iterations
are computed in FP32 for numerical stability, after which the mapped
residual streams are returned to the model activation dtype.

Defining the post-mapping logits as

\begin{equation}
    \mathbf{Z}_{\mathrm{post}}
    =
    \alpha_{\mathrm{post}}\mathbf{p}_{\mathrm{post}}
    +\mathbf{b}_{\mathrm{post}},
\end{equation}

the original mHC post-mapping is

\begin{equation}
    \mathbf{H}^{\mathrm{original}}_{\mathrm{post}}
    =
    2\sigma
    \left(
        \mathbf{Z}_{\mathrm{post}}
    \right).
    \label{eq:mhc_post_mapping}
\end{equation}

Motif 3 instead uses a time-dependent scale $s_t$,

\begin{equation}
    \mathbf{H}^{(t)}_{\mathrm{post}}
    =
    s_t\sigma
    \left(
        \mathbf{Z}_{\mathrm{post}}
    \right),
    \qquad
    s_t: 2 \longrightarrow 1,
    \label{eq:modified_mhc_post_mapping}
\end{equation}

which is gradually reduced from the original multiplier of two to one
during pretraining. With the original parameterization, logits near
zero produce an identity-scale post-mapping, but mapping values above
one can repeatedly amplify sublayer outputs as they propagate through
successive residual blocks. At the scale and depth of Motif 3, we found
that this repeated amplification led to a progressive accumulation of
large activation outliers.

We therefore retain the original scale at the beginning of training
and anneal $s_t$ from two to one. The post-mapping consequently
transitions from the range $(0,2)$ to $(0,1)$, preserving the original
early optimization behavior while removing persistent amplification
later in training. This modification limits the layer-wise accumulation
of activation outliers without applying hard clipping to the forward
activations.

\subsection{Expert-Specific PolyNorm}
\label{sec:expert_specific_polynorm}

Each sparse MoE layer contains 384 routed experts and one shared
expert, with eight routed experts selected for each token.
Because routing exposes different experts to different token
distributions, applying the same fixed activation to every expert
constrains them to share a common nonlinear response. Motif 3 instead
replaces the SiLU gate activation with Expert-Specific PolyNorm, whose
polynomial coefficients and bias are learned independently for each
expert. This allows each expert to adapt its activation shape to the
distribution of tokens routed to it.
For the tokens dispatched to expert $i$, the activation is

\begin{equation}
    \operatorname{PolyNorm}_{i}(\mathbf{z})
    =
    \sum_{n=1}^{3}
    a_{i,n}
    \frac{
        \mathbf{z}^{n}
    }{
        \operatorname{RMS}(\mathbf{z}^{n})
    }
    +
    b_i,
    \label{eq:polynorm}
\end{equation}

where the powers are applied element-wise and
$\operatorname{RMS}(\mathbf{u})
=\sqrt{d^{-1}\sum_{j=1}^{d}u_j^2+\epsilon}$.
To prevent the PolyNorm output from growing excessively, we introduce
a sigmoid parameterization for each polynomial coefficient:
$a_{i,n}=\sigma(\widetilde{a}_{i,n})$, which constrains
them to $(0,1)$, while the bias is constrained by
$b_i=\operatorname{clip}(\widetilde{b}_i,-0.5,0.5)$.
The operation is applied to the expert-grouped token buffer used by the MoE implementation; normalization is computed per token over the hidden dimension and is therefore unaffected by the grouping, with only the polynomial coefficients and bias differing across experts.
PolyNorm forms a learned linear combination of individually
RMS-normalized polynomial components, thereby decoupling each
component's magnitude from the scale of its input.
The expert-dependent coefficients promote specialization by allowing
different nonlinear responses across experts, while the normalized
polynomial basis and constrained parameters limit activation outliers.
In controlled experiments using models with approximately 10 billion
parameters, we quantify the diversity of the learned gate-weight
directions by measuring the effective rank of a gate matrix $\mathbf{W}$
as

\begin{equation}
    r_{\mathrm{eff}}(\mathbf{W})
    =
    \frac{
        \sum_j \sigma_j(\mathbf{W})^2
    }{
        \max_j \sigma_j(\mathbf{W})^2
    },
    \label{eq:gate_effective_rank}
\end{equation}

where $\sigma_j(\mathbf{W})$ are its singular values.
This quantity is larger when the singular-value energy is distributed
more evenly across multiple directions and smaller when the spectrum is
dominated by a single direction.
Equivalently, it is the sum of the normalized terms
$(\sigma_j/\sigma_{\max})^2$. Each term is at most one, and the number
of nonzero terms is $\operatorname{rank}(\mathbf{W})$; hence
$r_{\mathrm{eff}}(\mathbf{W})\leq\operatorname{rank}(\mathbf{W})$,
with equality when all nonzero singular values are equal to
$\sigma_{\max}$.
For the gate-weight matrices measured here,
$\operatorname{rank}(\mathbf{W})=512$, so the maximum possible effective
rank is 512.
Expert-Specific PolyNorm preserves a higher effective rank in the expert gate
weights than SwiGLU across the measured layers
(Figure~\ref{fig:gate_effective_rank}), indicating that the
learned gating transformations retain a broader set of active
directions.

Table~\ref{tab:architecture_summary} summarizes the principal model
dimensions and architectural configuration.

\begin{table}[t]
    \centering
    \small
    \renewcommand{\arraystretch}{1.08}
    \begin{tabular}{p{0.34\linewidth}p{0.57\linewidth}}
        \toprule
        \textbf{Property} & \textbf{Configuration} \\
        \midrule
        Total parameters
        & Approximately 314B \\
        Activated parameters
        & Approximately 13.2B per token \\
        Transformer layers
        & 53 (2 dense and 51 MoE) \\
        Hidden dimension
        & 4,096 \\
        Attention mechanism
        & Grouped Differential Latent Attention with output gating \\
        Attention pattern
        & Hybrid GDLA (1 full / 3 sliding-window) \\
        Query / KV heads
        & 80 / 16 \\
        Signal / noise query heads
        & 64 / 16 \\
        Query-key / Value head dim
        & 192 / 128 \\
        FFN dimension (dense / expert)
        & 12,288 / 1,280 \\
        Routed experts
        & 384, with top-8 routing \\
        Shared experts
        & 1 \\
        Expert activation
        & Expert-Specific PolyNorm \\
        Residual architecture
        & Modified manifold-constrained hyper-connections \\
        MTP head
        & 1 layer for self-speculative decoding \\
        Maximum context length
        & 262,144 tokens (256K) \\
        \bottomrule
    \end{tabular}
    \caption{Summary of the principal architectural configuration and
    model dimensions used in Motif 3, including its hybrid attention
    pattern, fine-grained expert structure, and long-context support.}
    \label{tab:architecture_summary}
\end{table}

\subsection{Multi-Token Prediction}
\label{sec:mtp}

We adopt the standard multi-token prediction layer introduced in
DeepSeek-V3~\cite{deepseekv3} as an auxiliary pretraining objective.
The resulting draft head can be used for self-speculative decoding. It
does not modify the primary autoregressive decoding architecture.

\section{Training System Optimization}

Reference implementations of the optimizations described in this section are available in the Motif~3 training example repository~\cite{motif3trainingexample}.

\subsection{Distributed Training Strategy}
\label{sec:distributed_training_strategy}

We train Motif~3 with a hierarchical parallel layout that follows the physical network topology. Expert parallelism (EP) is the innermost dimension with EP${}=8$, placing each expert-parallel group within a single eight-GPU node so that expert dispatch and combine communication use the high-bandwidth NVLink fabric. We use HybridEP~\cite{deepseekaihybridep} to implement the expert dispatch and combine operations within each EP group. Outside the expert-parallel dimension, we use a sharded data-parallel (DP-shard) degree of 8 and assign the remaining ranks to the replicated data-parallel (DP-replicate) dimension. With expert and parameter sharding, optimizer-state CPU offloading, and the memory capacity of B200 GPUs, our 314B-parameter model fits within the available device memory, and we therefore do not use pipeline parallelism (PP). To reduce communication overhead in this configuration, we apply the FSDP optimizations described below, including overlapping gradient Reduce-Scatter, suppressing redundant DP-replicate synchronization across accumulated microbatches, and eliminating unnecessary expert-weight All-Gathers. Context parallelism (CP), also referred to as sequence parallelism, is enabled only for long-context training, where CP${}=8$ reuses the same process dimension as EP, as described in Section~\ref{sec:long_context_training}.

\paragraph*{Activation Recomputation} Throughout pre-training and for supervised fine-tuning (SFT) at sequence lengths up to 128K, we use operator-level selective activation recomputation. Each transformer block is wrapped in a checkpoint region, and its forward pass is re-executed during backward according to a per-operator policy. Most intermediates are rematerialized, while a small set of outputs from the original forward is retained, including reduction outputs used to derive low-precision scaling factors.

The largest activations occur on the MoE path, where expert dispatch expands the token representation by the routing fan-out. We control this footprint through two coupled optimizations. First, we adopt the memory-efficient permutation strategy introduced in Megatron-Core~\cite{nvidia2026megatroncore}. Conventionally, each routing weight $p_i$ is applied after the expert down-projection as $p_i \mathbf{W}_2^{(i)}\mathbf{h}_i$, requiring the full expert output to remain available for computing the router-weight gradient. Because our expert down-projections are bias-free linear maps, we instead fold the routing weight into the permuted activation and compute the mathematically equivalent $\mathbf{W}_2^{(i)}(p_i\mathbf{h}_i)$. The router-weight gradient can then be obtained from the expert activation, and the inverse permutation reduces to an unweighted sum of routed contributions. Second, we expose this fused unpermute-and-combine operation to the selective checkpointing policy and retain its output. The post-combine tensor is smaller than its top-$k$-expanded inputs by approximately the routing fan-out, so retaining it has a modest memory cost while ensuring that backward recomputation neither retains the expanded expert activations nor replays the combine all-to-all. The dispatch all-to-all is re-executed to rematerialize expert inputs, but reuses the communication state and routing metadata cached during the original forward, avoiding additional host synchronization. Together, these measures reduce peak activation memory while preserving the model output.

At a context length of 256K tokens, activation memory dominates even under selective recomputation. We therefore apply full activation recomputation to every transformer block during SFT at this length, trading additional computation for the minimum activation footprint. Section~\ref{sec:long_context_training} describes the corresponding context-parallel execution strategy.

\paragraph*{Memory-Efficient Output Projection and Loss} With a vocabulary of approximately 220K entries, materializing the output logits would require more than 13~GB per microbatch in \texttt{BF16} even at the pre-training sequence length, with proportionally greater memory use at longer contexts. We instead compute the output projection and cross-entropy jointly using the Liger fused linear cross-entropy kernel~\cite{hsu2024liger}. The kernel processes hidden states in chunks, so the full logits tensor is never materialized during either the forward or backward pass. It returns an unnormalized token-level loss sum, which we normalize by the global number of valid tokens aggregated across data-parallel ranks and gradient-accumulation microbatches. This makes the loss and gradient scale invariant to the partitioning of each batch.

\paragraph*{Overlapping Gradient Reduce-Scatter} Each MoE block contributes up to three FSDP parameter groups: the dense-block parameters, the MoE non-expert parameters, and the routed experts. The routed experts are sharded over a separate device mesh. Consequently, gradient Reduce-Scatter operations are issued frequently during backward. The default FSDP schedule tracks a single in-flight Reduce-Scatter across these groups, forcing a group entering its post-backward phase to wait for the preceding operation before launching its own. This serializes reductions from expert and dense groups and repeatedly stalls backward computation. We modify FSDP to maintain a bounded window of in-flight Reduce-Scatter operations. Post-backward hooks enqueue reductions without immediately waiting, and their completion events are drained once at the end of the backward pass. This restores overlap between gradient reduction and backward computation while bounding the staging-buffer memory consumed by outstanding operations.

\paragraph*{Gradient Synchronization across Microbatches} We schedule gradient synchronization hierarchically over each gradient-accumulation cycle. Every microbatch Reduce-Scatters its gradients across the DP-shard dimension, causing gradients to accumulate in sharded \texttt{FP32} buffers rather than as full unsharded tensors. The All-Reduce across the DP-replicate dimension is suppressed for intermediate microbatches and issued only for the final microbatch over the already accumulated gradient shards. This reduces inter-replica gradient traffic by the gradient-accumulation factor while retaining the sharded per-microbatch memory footprint. The low-precision communication mechanism used for these reductions is described in the following subsection.

\paragraph*{Eliminating Expert Weight All-Gathers} FSDP normally hides parameter All-Gathers by prefetching them concurrently with computation. This overlap is not free, however, because communication kernels consume SM resources, memory capacity, and interconnect bandwidth, and can therefore slow the computation with which they overlap. Under the default policy of resharding parameters after forward and backward, expert weights would be gathered twice per MoE layer and microbatch. At our scale, we find it more efficient to eliminate these collectives than merely to hide them. During gradient accumulation, expert parameters are not resharded after backward on intermediate microbatches, allowing the gathered copy from one microbatch to be reused by the next; resharding is deferred until the final microbatch. This retention does not increase peak device memory in our configuration. As backward progresses, the saved activations associated with each checkpointed layer are released after that layer's gradients have been computed. The retained MXFP8 expert weights are smaller than the activations released over the same schedule, so the resident parameter copies occupy memory freed by backward rather than increasing the peak. For a selected subset of MoE layers (22 of 51 in the main pre-training configuration, chosen to satisfy the per-GPU memory budget), the experts additionally remain unsharded between forward and backward, reducing their weight gathering to exactly one All-Gather per optimizer step. Dense parameters retain the standard resharding policy.

Keeping resident unsharded expert copies is practical because expert weights are gathered in \texttt{MXFP8}. The \texttt{FP32} master-weight shards are quantized on the sender side, halving the All-Gather volume relative to \texttt{BF16}, and the gathered copies remain in \texttt{MXFP8} for direct use by the grouped FP8 GEMMs without dequantization. An unsharded resident copy therefore requires approximately half the memory of its \texttt{BF16} counterpart.

\subsection{Low-Precision Training}

We employ low-precision training to improve the efficiency of large-scale MoE training while preserving optimization stability. Although reduced-precision computation is now standard practice in modern large-model training, stable deployment still requires careful decisions about which states can be quantized aggressively and which should remain in high precision.

In our training setup, optimization-critical states, including the master weights, main gradients, router logits, and Muon optimizer states, are maintained in \texttt{FP32}. Low-precision formats are then applied selectively to the parts of training that dominate memory use, communication cost, and computational overhead.

\begin{figure*}[t]
    \centering
    \includegraphics[width=\textwidth]{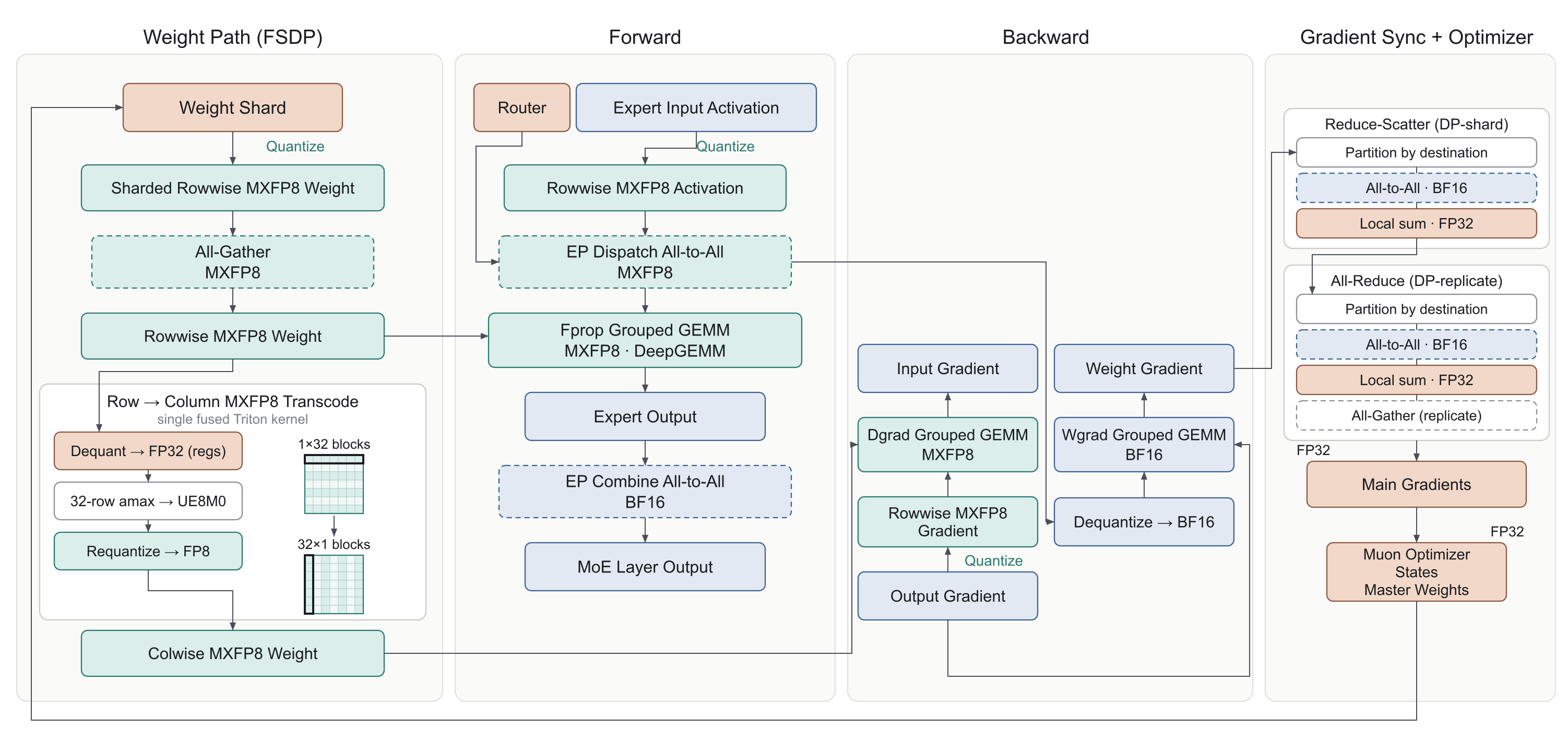}
    \caption{\small Overall low-precision training recipe. Color encodes numerical precision: teal for MXFP8, blue for BF16, and peach for FP32. Dashed borders mark collectives that cross ranks. Only the row-wise MXFP8 weight is All-Gathered; the column-wise copy required by Dgrad is produced locally by a fused row-to-column transcode kernel. Expert activations are quantized once before EP dispatch, allowing the dispatch itself to use MXFP8. Gradient synchronization exchanges BF16 shards while performing each reduction locally in FP32.}
    \label{fig:low_precision_overall_process}
\end{figure*}

More specifically, we apply MXFP8~\cite{rouhani2023microscaling} to the MoE experts, where a substantial fraction of expert-side computation is concentrated. This selective use of MXFP8 improves memory efficiency, reduces communication overhead, and increases computational throughput, while preserving the high-precision states required for stable optimization. Figure~\ref{fig:low_precision_overall_process} summarizes the numerical formats and communication paths used for expert weights, activations, gradients, and optimizer states. In the following paragraphs, we describe the specific low-precision design choices and system optimizations in more detail.

\paragraph*{MXFP8 Grouped GEMM} MoE computation accounts for a substantial fraction of end-to-end training cost, so applying low-precision arithmetic to the experts provides a meaningful training speedup. In our system, this optimization is centered on grouped GEMM for expert computation, where efficient MXFP8 execution is particularly important. Among several open-source MXFP8 GEMM implementations, including TransformerEngine~\cite{mishra2025mxfp8} and torchAO~\cite{or2025torchao}, we found DeepGEMM\footnote{\url{https://github.com/deepseek-ai/DeepGEMM}} to provide the best performance for our model scale on Blackwell GPUs, and therefore use it as our primary grouped GEMM backend.

\paragraph*{MoE Activation Pre-Quantization} Since MXFP8 Grouped GEMM takes MXFP8 tensors as input, the expert activations must also be quantized to \texttt{MXFP8}. Although this quantization could be applied immediately before Grouped GEMM, we instead quantize activations before EP dispatch so that the dispatch communication itself also benefits from reduced precision. This pre-quantization strategy does not introduce additional precision loss, while reducing the communication cost of expert dispatch by half. It also reduces quantization cost by a factor of top-$k$, since quantization is performed once before dispatch rather than separately for each selected expert; in our setup with top-$k=8$, this reduces the quantization cost to one eighth. In contrast, we do not explicitly downcast the output of MXFP8 Grouped GEMM from \texttt{BF16} to \texttt{MXFP8} for the combine path, since the GEMM output is already produced in \texttt{BF16} and forcing an additional downcast would add unnecessary quantization. To support the MXFP8 dispatch path, we make minor modifications to the HybridEP implementation so that it can natively handle \texttt{MXFP8} tensors during dispatch.

\paragraph*{Low-Precision for Gradient Synchronization} We use \texttt{BF16} for gradient synchronization across Fully Sharded Data Parallel (FSDP) ranks to reduce communication cost. Synchronization consists of Reduce-Scatter across DP-shard ranks and All-Reduce across DP-replicate ranks. Since the main gradients are maintained in \texttt{FP32}, conventional collectives would communicate them in \texttt{FP32}. We instead transfer gradient shards in \texttt{BF16} while performing the numerically sensitive reductions locally in \texttt{FP32}. For Reduce-Scatter, each rank partitions its gradient by destination, exchanges the \texttt{BF16} shards through all-to-all, and sums the received shards locally in \texttt{FP32}. For All-Reduce, we apply the same sharded reduction and then replicate the reduced shards across ranks with All-Gather. This reduces communication cost while maintaining stable distributed optimization.

\paragraph*{Row-to-Column MXFP8 Conversion} During the backward pass, both row-wise and column-wise MXFP8 weights are required for forward recomputation and gradient computation. Under FSDP, this would normally require All-Gather for both row-wise and column-wise MXFP8 tensors. Instead, we perform All-Gather only for the row-wise MXFP8 tensor and construct the column-wise tensor from the gathered representation, which reduces communication cost by half. Because row-wise and column-wise MXFP8 use different scaling axes, this conversion requires requantization and can introduce additional rounding error. We observe that this error does not measurably affect training stability or final model quality.

\subsection{Muon Optimizer}

We use Muon as the primary optimizer throughout Motif~3 training~\cite{jordan2024muon}. For each matrix-shaped parameter, Muon orthogonalizes the momentum update using a Newton--Schulz iteration. Embeddings, the output projection, and vector-shaped parameters are instead optimized with AdamW. Applying Muon at this scale introduces three systems challenges: orthogonalizing sharded parameters without redundant computation, monitoring attention logits cheaply enough to control their growth, and keeping optimizer states within device memory. We address these challenges in turn.

\paragraph*{Parallel Muon for Mixture-of-Experts} We compute the updates using the Parallel Muon algorithm introduced in our previous report~\cite{lim2025motif2}. Unlike Distributed Muon, which reconstructs each full matrix and redundantly executes its Newton--Schulz iteration on every participating rank~\cite{liu2025muon}, Parallel Muon assigns each parameter to a single owner rank. Parameters are sorted by the computational cost of orthogonalization and assigned to ranks in round-robin order to balance the workload. Gradient shards are gathered to the owner through an all-to-all, orthogonalized there, and redistributed as update shards through a second all-to-all. We process parameters in chunks and pipeline the gather, computation, and scatter phases, allowing the communication for one chunk to overlap the orthogonalization of another. Both the communication and the Newton--Schulz iteration use \texttt{BF16}.

For Motif~3, we extend this approach to the MoE parameters. Expert weights are partitioned along the expert dimension by expert parallelism and expert data-parallel sharding, so each rank already holds complete matrices for its local subset of experts and requires neither the gather nor the scatter phase. Each expert is an independent Muon orthogonalization unit whose update is normalized according to the shape of its weight matrix. We stack all locally held expert weights and apply the Newton--Schulz iteration to the entire batch using batched GEMMs, rather than invoking a two-dimensional kernel separately for each expert. Expert updates therefore require no optimizer communication, while the partitioning of experts distributes the computation across ranks. We additionally overlap this batched expert computation with the first all-to-all gather of the dense-parameter pipeline, placing the dense communication in flight while the expert updates are computed.

\paragraph*{QK-Clip} To control the attention-logit growth observed under Muon, we apply QK-Clip following MuonClip~\cite{kimiteam2025k2}. After an optimizer update, any attention head whose maximum pre-softmax logit $S_{\max}^{h}$ exceeds a threshold $\tau$ has its query and key projections rescaled to bring the logit magnitude back toward $\tau$, without otherwise changing the forward computation. We define $\gamma_h = \tau/S_{\max}^{h}$ for a head that requires clipping. For the MLA projections, we scale only the query rows, including the non-rotary and rotary components, and the non-rotary key rows. The value rows are never modified, and the decoupled rotary key, which is shared across heads, is left unchanged.

We account for the grouped-query attention structure when dividing the clipping burden between the query and key projections. A conventional equal split scales both sides by $\sqrt{\gamma_h}$. In our preliminary experiments, this rule repeatedly shrank the shared key projection faster than optimization could restore it, eventually causing the key weights to degenerate toward zero. Let $G$ denote the number of query heads that share one KV head. Because the shared key receives gradient contributions from $G$ query heads whose directions partially cancel, its effective recovery signal scales approximately as $1/\sqrt{G}$, whereas each query head retains its full recovery signal. We therefore reduce the clipping burden assigned to the shared key by setting the key-side clipping ratio to

\begin{equation}
    r
    =
    \frac{1}{1+\sqrt{G}},
    \label{eq:qk_clip_gqa_ratio}
\end{equation}

and scale the non-rotary query and key rows by $\gamma_h^{1-r}$ and $\gamma_h^r$, respectively. Since the shared rotary key is left unchanged, the rotary query rows are instead scaled by the full factor $\gamma_h$. Every logit component is therefore attenuated by exactly $\gamma_h$, while the asymmetric split prevents repeated clipping from collapsing the shared key weights. When multiple query heads share a key projection, we use the smallest $\gamma_h$ within the group, corresponding to the strongest required clipping, for the shared key rows.

The main challenge is monitoring $S_{\max}^{h}$. This value is the maximum over the pre-softmax attention-score matrix, which FlashAttention does not materialize, while explicitly recomputing $\mathbf{Q}\mathbf{K}^{\top}$ would add a quadratic pass at every layer. We extend the FlashAttention-4 (FA4) forward kernel~\cite{zadouri2026flashattention4} to emit the maximum for each head directly. Specifically, the running row maxima already maintained by the online softmax are reduced to a single scalar per head, adding negligible work to the attention forward pass. These values are max-reduced across microbatches and data-parallel ranks and supplied per layer to the optimizer, which rescales its local weight shards immediately after the parameter update. We perform monitoring and clipping every 10 steps, using $\tau=100$ during main pre-training and $\tau=200$ during the 256K-token long-context stage.

\paragraph*{Optimizer-State CPU Offloading} Muon maintains a single momentum buffer for each parameter. We offload these states to host memory after every optimizer step and reload them to the device only when they are next needed. All state tensors are packed into flat, pinned host buffers and transferred per tensor on dedicated streams, avoiding transient device-side staging buffers during both offload and reload.

A naive schedule places the host-to-device reload on the critical path immediately before the optimizer step. Instead, we initiate reloads from the activation-checkpoint post-forward hook at each layer boundary during backward, after the layer's parameter All-Gather completes and before its backward GEMMs begin. Expert optimizer states are therefore reloaded layer by layer in step with backward computation, allowing the transfers to overlap the backward GEMMs; the remaining states are reloaded in bulk before the optimizer step. A single hook site covers both the outer activation-checkpointed block and the inner expert FSDP group, and the save/load ordering prevents a checkpoint save from triggering a redundant offload--reload cycle.

\subsection{Kernel Optimization}

Several operators on the critical path are memory-bandwidth-bound or dominated by kernel-launch overhead in their eager form. We fuse these operators into single custom kernels, which removes intermediate round-trips through device memory and, where relevant, keeps the operators inside the compiled graph.

\paragraph*{Pad-Aware Fused GroupedPolyNorm} On the MoE expert path, each expert applies a PolyNorm activation before the down-projection. In eager execution this is a chain of elementwise and reduction operators, each launching its own kernel and materializing its own intermediate. Because experts run as a grouped GEMM, the activation operates on a token buffer that has been padded per expert to the grouped-GEMM alignment multiple, and those padded rows must not contribute to any per-group statistic. We implement a single fused kernel that computes the entire multiply and PolyNorm chain in one pass over the grouped, padded buffer, and is pad-aware: it consumes the per-expert group offsets and lengths so that padded rows are neither read into the normalization statistics nor written back with spurious values. The kernel provides both the forward and the closed-form backward, and is exposed as a drop-in replacement for the eager module so that tensor-parallel and expert-parallel sharding and the checkpoint key layout are unchanged.

\paragraph*{mHC Kernels} The mHC layer's per-step ``apply'' operations are inefficient in their natural matmul form. We replace them with elementwise multiply and multiply-sum formulations together with a Triton residual kernel, and implement the resulting layer to support full-graph \texttt{torch.compile}. The Sinkhorn-Knopp normalization and the associated projections are computed in \texttt{FP32} for numerical stability.

\paragraph*{Fused PolyNorm--FP8 Inference Kernel} For inference we use a fused kernel that applies PolyNorm and FP8 quantization together. Decode is strongly memory-bandwidth-bound, so fusing the normalization with the quantization avoids a separate read and write of the hidden state to produce the FP8 operand and yields a disproportionately large benefit at inference relative to the same fusion during training.

\subsection{Long-Context Training}
\label{sec:long_context_training}

We extend training to sequences of up to 256K tokens using EP${}=8$ and CP${}=8$, with the context-parallel group reusing the same eight-rank process dimension as the expert-parallel group. As described in Section~\ref{sec:distributed_training_strategy}, this stage uses full activation recomputation to control its activation-memory footprint. At this context length, no single context-parallel algorithm is efficient for both full- and sliding-window-attention layers, while document-masked packing creates substantial workload variation across nominally equal-length sequences. We address these challenges with window-aware context parallelization within each sequence and attention-workload-aware re-scheduling across data-parallel ranks and gradient-accumulation microbatches. Throughout this subsection, $L$ denotes the packed sequence length, $W$ the sliding-window size, and $P$ the context-parallel degree. We use $h_q$, $h_k$, $h_v$, and $h_o$ for the numbers of query, key, value, and output heads, and $d_{qk}$ and $d_v$ for the query/key and value head dimensions.

%
\begin{figure}[t]
  \centering
  \resizebox{\linewidth}{!}{%
    \input{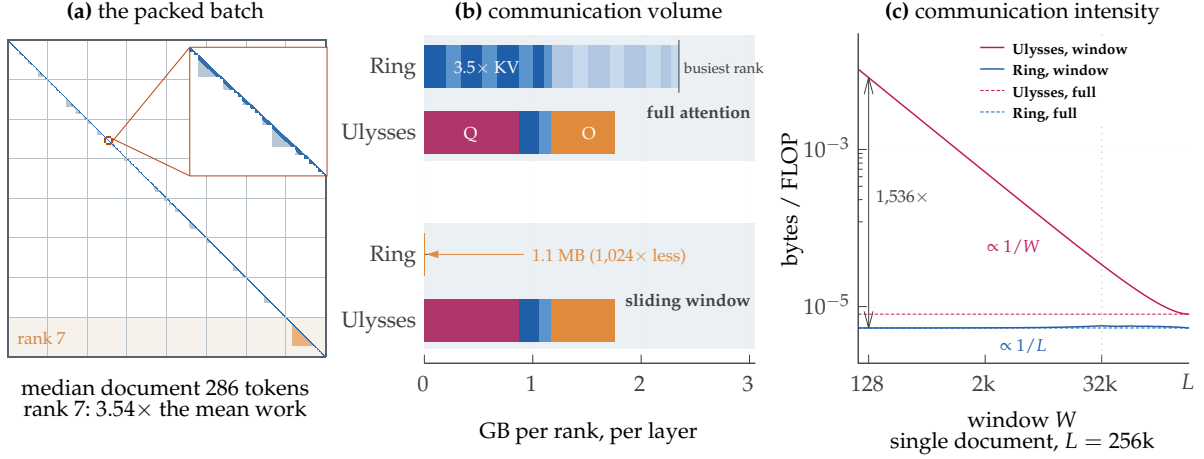}%
  }
  \caption{%
    \textbf{Selecting the context-parallel algorithm per attention layer.}
    \textbf{(a)}~A packed sequence sampled from the training data and
    containing $473$ documents, shown on the attention score matrix
    $QK^{\top}$.
    \textbf{(b)}~Rank-average forward-pass communication volume per layer. The
    marker above the Ring Attention bar reports the final rank, which receives
    twice the rank average.
    \textbf{(c)}~Rank-average bytes sent per attention FLOP versus the window
    size~$W$, evaluated on a single document.
    All panels use $P=8$, $L=256$K tokens, the head configuration of Section~\ref{sec:pretraining_setups}, and BF16.
    Panels \textbf{(a)} and \textbf{(b)} use $W=128$
    tokens.%
  }
  \label{fig:window-aware-sp}
\end{figure}

\paragraph*{Window-Aware Context Parallelization} Ring Attention~\cite{liu2023ringattention} is a common implementation of CP: each rank owns a query block and circulates key and value blocks through the CP group. When document-aware attention kernels skip masked cross-document blocks, however, the work assigned to a rank depends on how packed-document boundaries intersect its local query block. Variable document lengths can therefore produce substantial rank-level imbalance (Figure~\ref{fig:window-aware-sp}a). Static Striped Attention and zigzag Ring Attention layouts balance the triangular workload induced by a causal mask~\cite{brandon2023striped,zhu2024zigzag}, but do not in general balance document-masked packing because the active attention blocks depend on the input-specific document boundaries. Dynamic Context Parallelism instead uses fine-grained, input-dependent partitioning and device assignment~\cite{jiang2025dcp}, at the cost of scheduling and remapping attention blocks at each iteration.

Ulysses~\cite{jacobs2023ulysses} uses all-to-all communication to transpose the sequence and attention-head dimensions. When the query and key--value head counts are divisible by $P$, each rank processes the complete packed sequence for an equal subset of heads, removing this source of rank-level imbalance; this condition holds in our configuration because $h_q=80$, $h_k=16$, and $P=8$. In the implementations we considered, including ours, the data dependency between this transpose and attention prevents overlapping the all-to-all with attention computation. Under sliding-window attention, the attention computation drops from $\Theta(L^2)$ to $\Theta(LW)$, while the Ulysses all-to-all volume remains $\Theta(L)$. Applying Ulysses to these layers therefore yields an unfavorable communication-to-computation ratio for small $W$ (Figures~\ref{fig:window-aware-sp}b and~\ref{fig:window-aware-sp}c).

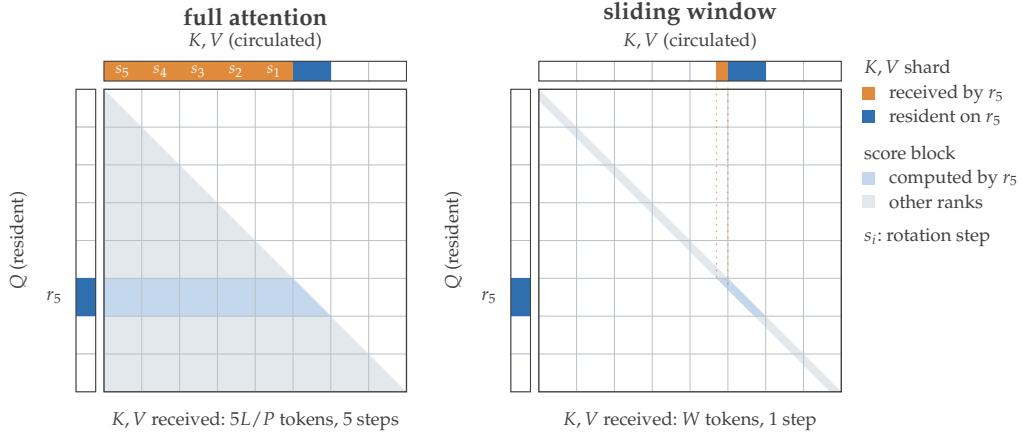
\begin{figure}[t]
  \centering
  \begin{tikzpicture}[x=1cm, y=1cm, font=\footnotesize]
    \definecolor{cOwn}{HTML}{2f6fb0}    
    \definecolor{cRecv}{HTML}{e08a3c}   
    \definecolor{cMine}{HTML}{c3d8ec}   
    \definecolor{cMask}{HTML}{e3e8ed}
    \definecolor{cInk}{HTML}{404040}
    \definecolor{cEdge}{HTML}{b9c1c8}

    \def\CS{0.50}      
    \def\TW{0.26}      
    \def\SP{0.11}      
    \def\ME{5}
    \def\HW{0.15}      
    \def\GAP{5.75}

    \newcommand{\drawmap}[2]{%
      \begin{scope}[xshift=#1cm]
        \ifnum#2=0
          \fill[cMask] (0,0) -- (8*\CS,-8*\CS) -- (0,-8*\CS) -- cycle;
          \fill[cMine] (0,-\ME*\CS) -- (\ME*\CS,-\ME*\CS)
                       -- (\ME*\CS+\CS,-\ME*\CS-\CS) -- (0,-\ME*\CS-\CS)
                       -- cycle;
        \else
          \fill[cMask] (0,0) -- (8*\CS,-8*\CS) -- (8*\CS-\HW,-8*\CS)
                       -- (0,-\HW) -- cycle;
          \fill[cMine] (\ME*\CS,-\ME*\CS) -- (\ME*\CS+\CS,-\ME*\CS-\CS)
                       -- (\ME*\CS+\CS-\HW,-\ME*\CS-\CS)
                       -- (\ME*\CS-\HW,-\ME*\CS) -- cycle;
        \fi
        \foreach \k in {0,...,8}{
          \draw[cEdge, line width=0.25pt] (\k*\CS,0) -- (\k*\CS,-8*\CS);
          \draw[cEdge, line width=0.25pt] (0,-\k*\CS) -- (8*\CS,-\k*\CS);
        }
        \draw[cInk, line width=0.5pt] (0,0) rectangle (8*\CS,-8*\CS);
        \ifnum#2=1
          \draw[cRecv, line width=0.45pt, dotted]
            (\ME*\CS-\HW,\SP) -- (\ME*\CS-\HW,-\ME*\CS);
          \draw[cRecv, line width=0.45pt, dotted]
            (\ME*\CS,\SP) -- (\ME*\CS,-\ME*\CS-\HW);
        \fi

        \begin{scope}[yshift=\SP cm]
          \foreach \j in {0,...,7}{
            \draw[fill=white, draw=cEdge, line width=0.25pt]
              (\j*\CS,0) rectangle (\j*\CS+\CS,\TW);
          }
          \ifnum#2=0
            \foreach \j [count=\s from 1] in {4,3,2,1,0}{
              \fill[cRecv] (\j*\CS,0) rectangle (\j*\CS+\CS,\TW);
              \node[white, font=\tiny] at (\j*\CS+\CS/2,\TW/2) {$s_\s$};
            }
            \fill[cOwn] (\ME*\CS,0) rectangle (\ME*\CS+\CS,\TW);
          \else
            \fill[cRecv] (\ME*\CS-\HW,0) rectangle (\ME*\CS,\TW);
            \fill[cOwn] (\ME*\CS,0) rectangle (\ME*\CS+\CS,\TW);
          \fi
          \draw[cInk, line width=0.4pt] (0,0) rectangle (8*\CS,\TW);
          \node[anchor=south, cInk, font=\scriptsize] at (4*\CS,\TW+0.03)
            {$K,V$ (circulated)};
        \end{scope}

        \begin{scope}[xshift=-\SP cm]
          \foreach \i in {0,...,7}{
            \draw[fill=white, draw=cEdge, line width=0.25pt]
              (-\TW,-\i*\CS) rectangle (0,-\i*\CS-\CS);
          }
          \fill[cOwn] (-\TW,-\ME*\CS) rectangle (0,-\ME*\CS-\CS);
          \draw[cInk, line width=0.4pt] (-\TW,0) rectangle (0,-8*\CS);
          \node[anchor=east, cInk, font=\scriptsize]
            at (-\TW-0.06,-\ME*\CS-\CS/2) {$r_5$};
          \node[rotate=90, anchor=south, cInk, font=\scriptsize]
            at (-\TW-0.52,-4*\CS) {$Q$ (resident)};
        \end{scope}
      \end{scope}}

    \drawmap{0}{0}
    \drawmap{\GAP}{1}

    \node[anchor=south, cInk] at (4*\CS,0.72) {\textbf{full attention}};
    \node[anchor=south, cInk] at (\GAP+4*\CS,0.72) {\textbf{sliding window}};
    \node[anchor=north, cInk, font=\scriptsize] at (4*\CS,-8*\CS-0.14)
      {$K,V$ received: $5L/P$ tokens, $5$ steps};
    \node[anchor=north, cInk, font=\scriptsize] at (\GAP+4*\CS,-8*\CS-0.14)
      {$K,V$ received: $W$ tokens, $1$ step};

    \begin{scope}[shift={(9.93,0.28)}]
      \node[anchor=west, cInk, font=\scriptsize] at (0,0) {$K,V$ shard};
      \foreach \c/\t [count=\k from 0] in {%
        cRecv/{received by $r_5$}, cOwn/{resident on $r_5$}}{
        \fill[\c] (0.10,-0.42-\k*0.30) rectangle (0.26,-0.42-\k*0.30+0.16);
        \node[anchor=west, cInk, font=\scriptsize]
          at (0.33,-0.42-\k*0.30+0.08) {\t};
      }
      \node[anchor=west, cInk, font=\scriptsize] at (0,-1.12) {score block};
      \foreach \c/\t [count=\k from 0] in {%
        cMine/{computed by $r_5$}, cMask/{other ranks}}{
        \fill[\c] (0.10,-1.54-\k*0.30) rectangle (0.26,-1.54-\k*0.30+0.16);
        \node[anchor=west, cInk, font=\scriptsize]
          at (0.33,-1.54-\k*0.30+0.08) {\t};
      }
      \node[anchor=west, cInk, font=\scriptsize] at (0,-2.24)
        {$s_i$: rotation step};
    \end{scope}
  \end{tikzpicture}
  \caption{%
    \textbf{Ring Attention communication under full and sliding-window
    attention.}  The block structure of $QK^{\top}$ for $P=8$
    context-parallel ranks, shaded from the perspective of rank~$5$: query
    shards on the left, key/value shards on top, score blocks in between.
    Only the key and value shards are exchanged.  Full attention requires every preceding shard,
    one per rotation step; a window of $W \le L/P$ tokens requires only the
    trailing $W$ tokens of the preceding shard, reducing the per-rank volume
    from $\Theta(L)$ to $\Theta(W)$.  Exchanged tokens are drawn far wider
    than $W/L$ to remain visible; forwarding hops of the standard ring schedule are
    omitted.%
  }
  \label{fig:ring-window-collapse}
\end{figure}

We therefore select the context-parallel algorithm per layer: sliding-window layers use window-aware Ring Attention, while full-attention layers use Ulysses. For full attention, Ulysses removes the document-dependent rank imbalance of Figure~\ref{fig:window-aware-sp}a by construction, whereas retaining Ring Attention would require the per-iteration dynamic block scheduling discussed above. The cost of this choice, an all-to-all that is not overlapped with attention computation, is modest for two reasons: the $\Theta(L)$ communication is amortized by the $\Theta(L^2)$ attention computation of these layers, and because the context-parallel group shares its eight-GPU node with the expert-parallel group (Section~\ref{sec:distributed_training_strategy}), the all-to-all traverses only the intra-node NVLink fabric. For the sliding-window layers, with $W \le L/P$, each rank needs only the trailing $W$ tokens of the preceding rank's KV shard (Figure~\ref{fig:ring-window-collapse}). This single halo exchange replaces up to $P-1$ rotation steps and reduces the rank-average received KV volume from $\Theta(L)$ to $\Theta(W)$: an analytic factor of $L/(2W)=1{,}024\times$ at our 128-token operating window in Figure~\ref{fig:window-aware-sp}b. The fixed window also bounds rank-level workload variation, so these layers do not require input-dependent block scheduling.

Although a Ring rotation can in principle be hidden behind attention computation, this overlap presumes balanced per-step work: under document masking, ranks whose query blocks intersect few unmasked score blocks expose the communication, and each synchronous rotation step is gated by the slowest rank, the same imbalance shown in Figure~\ref{fig:window-aware-sp}a. In our long-context runs, full-attention layers under Ulysses did not become step-level stragglers, whereas document-masked Ring variants did.

The communication-intensity sweep in Figure~\ref{fig:window-aware-sp}c is evaluated on a single document,\footnote{\label{fn:causal-window-score-count}For a causal window over a single document of length $L$, the number of unmasked score entries is $\sum_q \min(W,q+1) = W(W+1)/2 + (L-W)W$, which reduces to $L(L+1)/2$ at $W=L$.} isolating the dependence on $W$ from document-packing effects; Appendix~\ref{app:cp-volume} gives the communication-volume accounting underlying Figure~\ref{fig:window-aware-sp}. The \emph{Ring, full} curve uses the causal early-exit accounting of Appendix~\ref{app:cp-volume}.\footnote{Beyond $W=L/P$ (dotted line in Figure~\ref{fig:window-aware-sp}c), the window-aware curve is an idealized lower bound: the halo then extends beyond the immediately preceding shard, so the implementation reverts to block-granular rotation and transfers more data than plotted.} On rank average, window-aware Ring Attention remains below the Ulysses volume for every window size: under causal masking, each rank requires only the key--value blocks of its preceding ranks, and the halo exchange delivers exactly this set, at most two-thirds of the Ulysses all-to-all even at $W=L$. At our 128-token operating window, the gap is $1{,}536\times$ (Figure~\ref{fig:window-aware-sp}c).

%
%
%

\paragraph*{Attention-Workload-Aware Re-Scheduling} Document-masked packing concatenates multiple documents into a fixed-length sequence while preventing cross-document attention. Although packed sequences have the same nominal length, their document boundaries can produce substantially different attention workloads. For example, in the packed sequence of Figure~\ref{fig:window-aware-sp}a, rank~$7$'s attention FLOPs are $3.54\times$ the cross-rank mean. In synchronous training, this variation creates step-level stragglers, meaning that each microbatch duration, and hence the step time, is bounded by the slowest data-parallel replica.

To quantify this variation, we use the number of unmasked attention-score entries as a proxy for attention FLOPs. For a document of length $d$, define the full-attention score count as $n_{\mathrm{full}}(d)=d(d+1)/2$ and the sliding-window score count as $n_W(d)=\sum_{q=0}^{d-1}\min(W,q+1)$ (closed form in footnote~\ref{fn:causal-window-score-count}). Because one quarter of our layers use full attention and three quarters use sliding-window attention, we estimate the work of a packed sequence with document lengths $\{d_i\}$ by
\begin{equation}
  \widehat{C}(\{d_i\})
  = \sum_i \left[\frac{1}{4}n_{\mathrm{full}}(d_i)
  + \frac{3}{4}n_W(d_i)\right].
\end{equation}
For equal nominal sequence lengths and fixed $W$, the sliding-window term is nearly linear in token count, whereas the quadratic full-attention term drives most of the variation across document layouts.

We rebalance complete packed sequences across the data-parallel dimension. Let $N_{\mathrm{DP}}$ be the data-parallel group size, $N_{\mathrm{acc}}$ the number of gradient-accumulation microbatches, and $B$ the local microbatch size. Each replica prefetches the sequences required for one optimizer step, producing a global pool of $N_{\mathrm{DP}}N_{\mathrm{acc}}B$ packed sequences; our configuration uses $B=1$. The replicas all-gather the per-sequence cost estimate $\widehat{C}$ defined above. Given this identical input, every replica independently runs the same deterministic longest-processing-time greedy heuristic: sequences are sorted by decreasing estimated cost and assigned in turn to the currently least-loaded replica--microbatch slot. This approximately minimizes the maximum estimated work per microbatch; deterministic tie-breaking makes every replica produce the same assignment without a separate broadcast. The sequences are then exchanged through a fixed-size all-to-all because all packed sequences share the same nominal length. This procedure is orthogonal to CP, which is applied within each data-parallel replica. Because the same set of sequences contributes to the optimizer step with the same weighting, the accumulated gradient is preserved up to floating-point reduction order; only the assignment to data-parallel replicas and microbatches changes.

\section{Pre-training}

\subsection{Data and Tokenizer}
\label{sec:pretraining_data}
\paragraph{Data.} 
Motif 3 is pretrained on approximately 12.5 trillion tokens, measured
with the Motif tokenizer, from a broad mixture of general, technical,
multilingual, and reasoning-oriented sources. General web documents
form the largest component of the corpus, supplemented by substantial
amounts of STEM material, source code, synthetic question-answer
pairs, and mathematical content.
We additionally include dedicated Korean and multilingual corpora,
reasoning-intensive examples, and domain-specific data from the legal
and financial sectors.
The broader mixture incorporates all publicly released components of
NVIDIA's Nemotron pretraining collection~\cite{nvidia2026nemotronpretraining}.
We apply additional filtering to
some components. The retained Nemotron data account for approximately
70\% of the full Motif 3 pretraining corpus.
Most of the remaining data, including web documents, STEM material,
and synthetic examples, were collected and processed in-house. The
latest data included in the corpus were collected through March 2026,
which we use as the model's knowledge cutoff.

The included releases comprise Nemotron-Pretraining-Dataset-sample,
Nemotron-Pretraining-Legal-v1, Nemotron-Pretraining-Specialized-v1,
v1.1, and v1.2, Nemotron-Pretraining-Code-v1, v2, and v3,
Nemotron-CC-v2 and v2.1, Nemotron-CC-Code-v1, Nemotron-CC-Math-v1, and
Nemotron-Pretraining-SFT-v1. Together, these sources provide broad web
text and targeted coverage of code, mathematics, legal material,
specialized domains, and instruction-like data. These public releases
form only part of the complete pretraining corpus, and the list above is
not an exhaustive account of the data used to train Motif 3.

This composition is designed to preserve broad linguistic and factual
coverage while strengthening the technical reasoning, code generation,
mathematical problem solving, and Korean-language capabilities of the
model.

The Motif tokenizer provides higher average compression than the
comparison tokenizers on the multilingual and multi-domain probe in
Table~\ref{tab:tokenizer_compression}. Considering the compression
rates on the English, Korean, and mathematics partitions, the
12.5-trillion-token training volume corresponds to more than 15
trillion tokens when expressed using any of the comparison tokenizers.
This is a probe-based estimate rather than a complete retokenization of
the pretraining corpus.

\paragraph{Dynamic mixture scheduling.}
Following Motif~2~\cite{lim2025motif2}, we use dynamic data-mixture
scheduling rather than a single fixed sampling distribution. Datasets
are first grouped into semantic categories corresponding to the major
components of the corpus. The initial mixture contains 19 constituent
datasets. As training progresses, we expand the pool to 57 datasets and
adjust their sampling weights through the dynamic mixture schedule. At
every training step, the scheduler converts the current category
ratios into integer sample allocations for the batch. Within each
category, samples are assigned across constituent datasets in
proportion to their available sample counts throughout training.

The category ratios remain at their initial values for a configurable
fixed phase and then move toward target ratios using either linear or
cosine interpolation. This enables general-domain data to dominate the
early phase while progressively increasing the contribution of STEM,
mathematics, code, synthetic QA, and reasoning-oriented sources. The
mixture is therefore adjusted continuously at the step level over the
full 12.5-trillion-token training run.

\paragraph{Tokenizer.} 
We trained the Motif~3 tokenizer from scratch using the SuperBPE reference
implementation\footnote{\texttt{https://github.com/pythonnut/superbpe}}~\cite{liu2025superbpe}.
Tokenizer training proceeds in two stages. Stage~1 learns a conventional BPE
vocabulary, while Stage~2 augments the pre-tokenization pattern with repeated
space-separated letter runs. This allows multiple words to be merged within a
single pre-tokenization unit. The complete patterns for both stages are provided in
Appendix~\ref{app:tokenizer_regexes}.
This enables the tokenizer to learn \emph{superword} tokens that span multiple
whitespace-delimited words. Examples include \textnormal{``of the''} in English and
\textnormal{``수 있다''} in Korean. The modified pattern preserves the original
handling of numbers, punctuation, line breaks, and whitespace. By encoding
frequent multi-word sequences with fewer tokens,
SuperBPE is particularly effective for whitespace-delimited languages such as
English and Korean.

\paragraph{Compression efficiency.}
We compare tokenizer compression against recent open-source tokenizers using
bytes per token, where a higher value indicates that fewer tokens are required
to encode the same text. Measurements are taken on a held-out multilingual,
multi-domain probe containing approximately 23~MB of text. The English,
French, Korean, Japanese, and Chinese partitions each contain 2,000,000
characters sampled from the November~1, 2023 snapshot of
\texttt{wikimedia/wikipedia}. The code and mathematics partitions each contain
2,000,000 characters sampled from the corresponding pretraining sources,
Nemotron-CC-Code-v1 and Nemotron-CC-Math-v1.

For each tokenizer, we divide the number of UTF-8 bytes by the number of
produced tokens, with special tokens disabled during encoding. Baseline
tokenizers are loaded from their official Hugging Face releases. The
vocabulary column in Table~\ref{tab:tokenizer_compression} reports the full
released vocabulary size, including any reserved or special tokens. Motif~3
achieves the strongest compression on English, Korean, code, and mathematics,
while remaining competitive on the other language partitions.

\begin{center}
    \centering
    \captionof{table}{Tokenizer compression measured in bytes per token. A
    higher value corresponds to fewer tokens for the same text. The best value
    in each partition is shown in bold.}
    \label{tab:tokenizer_compression}
    \resizebox{0.66\linewidth}{!}{%
    \begin{tabular}{lrrrrrrrr}
        \toprule
        Tokenizer & Vocab & en & fr & ko & ja & zh & code & math \\
        \midrule
        Motif (ours) & 220,160 & \textbf{5.68} & 4.02 & \textbf{5.31} & 3.87 & 3.51 & \textbf{4.07} & \textbf{4.55} \\
        Qwen3.5 & 248,066 & 4.51 & 3.91 & 4.03 & \textbf{4.22} & 4.13 & 3.68 & 3.54 \\
        Gemma-4 & 262,144 & 4.61 & 3.96 & 3.73 & 4.18 & 3.67 & 3.57 & 3.58 \\
        DeepSeek-V4 & 129,280 & 4.73 & 3.76 & 3.34 & 3.79 & \textbf{4.20} & 3.80 & 3.86 \\
        gpt-4o (o200k) & 200,000 & 4.70 & \textbf{4.14} & 3.65 & 3.43 & 3.31 & 3.92 & 3.75 \\
        \bottomrule
    \end{tabular}%
    }
\end{center}

\subsection{MoE Training Stabilization}
\label{sec:moe_stabilization}

A central challenge in training sparse MoE models is the positive
feedback loop between expert utilization and expert quality.
When the router initially favors a small subset of experts, those
experts receive more training tokens and stronger gradients, which can
make them increasingly likely to be selected.
Meanwhile, underutilized experts receive insufficient gradients and
may become effectively inactive.
Even when token counts appear balanced, experts may converge toward
similar functions and lose the specialization that motivates the MoE
architecture.
We therefore employ multiple complementary stabilization mechanisms
that address routing imbalance, expert starvation, specialization
collapse, and activation outliers.
The modified mHC and Expert-Specific PolyNorm architectural components are
described in Sections~\ref{sec:modified_mhc}
and~\ref{sec:expert_specific_polynorm}, respectively.

\paragraph{FP32 routing and selective load balancing.}

Motif 3 follows the established sigmoid-routing and auxiliary-loss-free
expert-selection bias design used in prior sparse MoE models
\cite{wang2024lossfree,deepseekv3}.
We keep the routing computation, including router scores and
expert-selection biases, entirely in FP32 to avoid numerical errors in
the discrete expert-selection decisions.

We additionally apply a sequence-wise auxiliary load-balancing loss
with sum reduction. During the early phase of pretraining, the earliest
MoE layers were more susceptible to expert-load imbalance than the
remaining layers, so we assign them stronger layer-specific auxiliary
loss weights. The bias-update coefficient and all auxiliary loss
weights are specified in the optimization hyperparameters below.
Despite this stronger balancing pressure, their routing balance still
collapsed occasionally.
When a collapse was detected, we immediately replaced the affected
layer's router parameters and expert-selection biases with those from
an adjacent layer that exhibited stable expert utilization.
Following these selective replacements, expert balance remained stable
and we observed no further routing collapse.

\paragraph{Decaying router noise.}

Routing decisions made during the early phase of training can become
self-reinforcing before experts have developed meaningful
specialization.
To encourage exploration, we perturb the FP32 router logits with
Gaussian noise:

\begin{equation}
    \widetilde{r}_i^{(t)}
    =
    r_i^{(t)}
    +
    \epsilon_i^{(t)},
    \qquad
    \epsilon_i^{(t)}
    \sim
    \mathcal{N}
    \left(
        0,
        \sigma_t^2
    \right).
    \label{eq:router_noise}
\end{equation}

\begin{figure*}[t]
    \centering
    \begin{subfigure}[t]{0.48\textwidth}
        \centering
        \includegraphics[width=\linewidth]{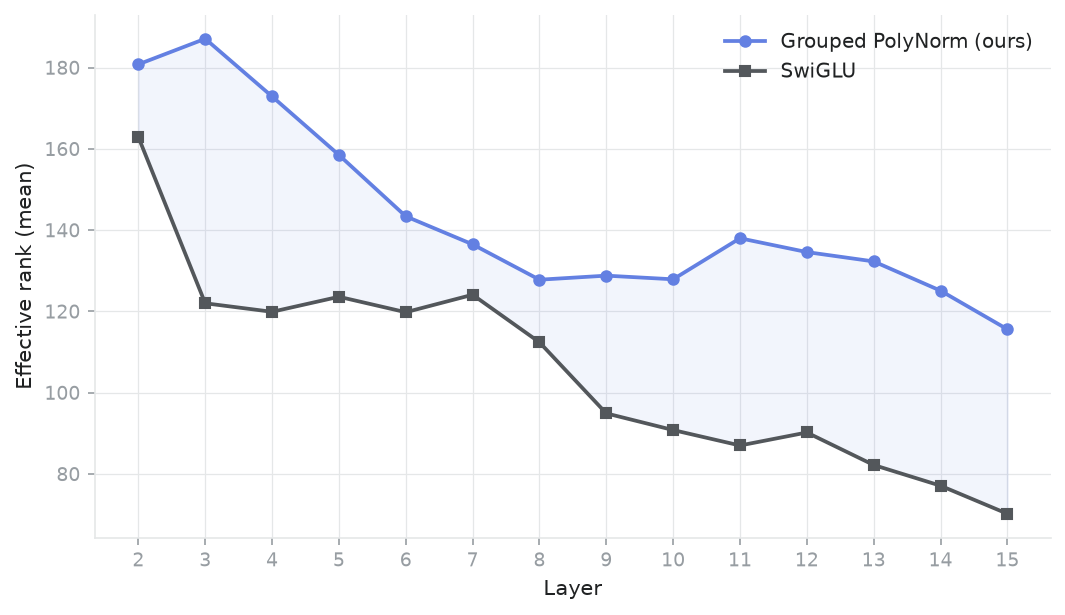}
        \caption{Gate-weight effective rank.}
        \label{fig:gate_effective_rank}
    \end{subfigure}
    \hfill
    \begin{subfigure}[t]{0.48\textwidth}
        \centering
        \includegraphics[width=\linewidth]{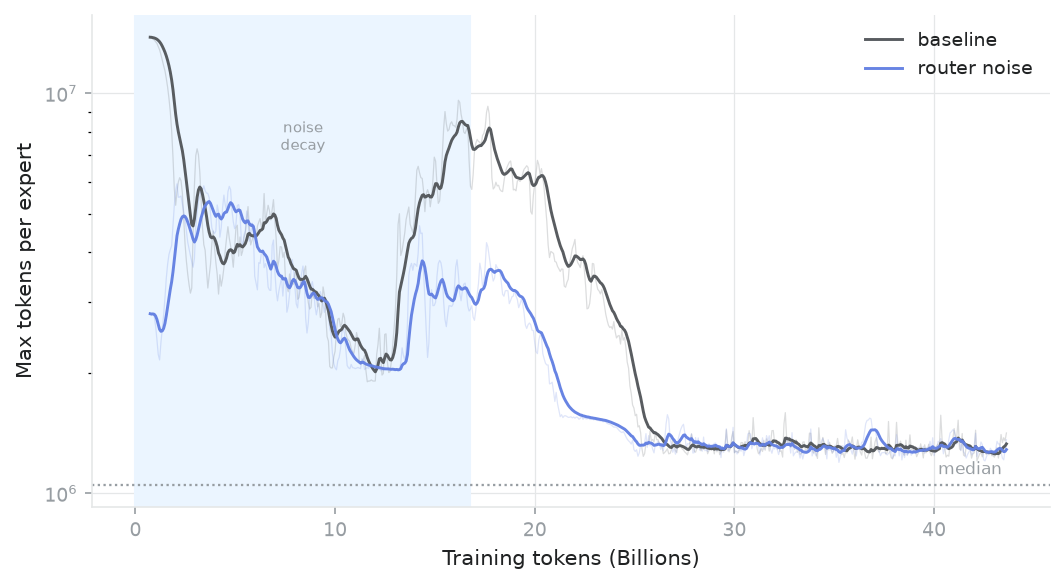}
        \caption{Maximum expert load.}
        \label{fig:router_noise_max_tokens}
    \end{subfigure}
    \caption{
        MoE component training comparisons from controlled experiments
        using models with approximately 10 billion parameters.
        \textbf{(a)} Expert-Specific PolyNorm maintains a higher mean effective
        rank in the expert gate weights than SwiGLU across layers,
        indicating a more evenly distributed singular-value spectrum.
        \textbf{(b)} Decaying router noise reduces the maximum number of
        tokens assigned to an expert more rapidly and guides the routing
        distribution toward the median-load regime early in training.
    }
    \label{fig:moe_component_analysis}
\end{figure*}

The noise scale follows a cosine decay schedule,

\begin{equation}
    \sigma_t
    =
    \sigma_{\min}
    +
    \frac{1}{2}
    \left(
        \sigma_{\mathrm{start}}-\sigma_{\min}
    \right)
    \left[
        1+
        \cos
        \left(
            \pi
            \min
            \left(
                \frac{t}{T_{\mathrm{noise}}},
                1
            \right)
        \right)
    \right],
    \label{eq:router_noise_decay}
\end{equation}

where $T_{\mathrm{noise}}$ denotes the router-noise decay period.
A relatively large initial noise scale exposes a wider range of
experts to training tokens and reduces the risk of premature routing
lock-in.
The noise is progressively removed after expert roles begin to form,
preventing continued stochastic exploration from disrupting mature
routing patterns.
In practice, this early-stage router noise substantially accelerates
the formation of balanced expert loads and causes the training loss to
decrease more rapidly during the initial phase of pretraining.
Figure~\ref{fig:router_noise_max_tokens} shows that the maximum token
load per expert falls toward the median-load regime substantially
earlier with router noise than in the baseline without noise.
These benefits are concentrated at the beginning of training; once
the routing distribution has stabilized, annealing the noise away
preserves the established balance without adding unnecessary
stochasticity to later optimization.

\paragraph{FFN magnitude regularization.}

Large activation outliers become increasingly common as language-model
training progresses.
In gated FFNs, the element-wise product between the activated gate and
the linear branch can further amplify extreme values, creating large
intermediate or output activations that reduce numerical headroom and
may precede training instability.
One common mitigation is to impose hard bounds inside SwiGLU.
For example, the reference implementation of gpt-oss-120b clamps the
gate input from above and clamps the linear input on both sides before
applying the gated product~\cite{openai2025gptossrepo}.

We avoid this forward-pass intervention because activation outliers are
not necessarily meaningless numerical artifacts.
Recent work shows that extreme activations can interact with
normalization to provide useful rescaling behavior, and that directly
clipping them can degrade both training stability and model
performance~\cite{qiu2026outlierrescaling}.
Hard clipping also discards magnitude information and eliminates
gradients once an activation crosses the clipping threshold.
Instead, we preserve the original forward activations and apply a soft
auxiliary magnitude penalty to each FFN output
$\mathbf{y}^{(\ell)}$:

\begin{equation}
    \mathcal{L}^{(\ell)}_{\mathrm{ffn\text{-}clip}}
    =
    w_{\mathrm{ffn}}
    \operatorname{mean}
    \left[
        \operatorname{ReLU}
        \left(
            |\mathbf{y}^{(\ell)}|-\tau
        \right)^2
    \right].
    \label{eq:ffn_clip_loss}
\end{equation}

The loss weight and activation thresholds are specified in the
optimization hyperparameters below.

This objective leaves normal activations below $\tau$ unchanged and
smoothly penalizes only excessive magnitudes.
It therefore discourages exploding FFN channels and
super-weight-like outliers while preserving the forward computation
and maintaining a nonzero recovery gradient for values above the
threshold.
After introducing this loss, we no longer observed isolated extreme
outliers.
However, the activation growth became distributed across a broader set
of channels, and the overall FFN output RMS continued to increase over
training.
The auxiliary loss therefore suppresses concentrated spikes but does
not by itself prevent global activation-scale drift, which we track
separately through the layer-wise RMS statistics described below.

\paragraph{Expert-health monitoring.}

The preceding balancing mechanisms are accompanied by continuous
layer-wise monitoring of router activity, expert utilization, and
expert-output statistics.
We collect these metrics at the same interval as the expert-utilization
heatmap and only on the final microbatch of the step to limit monitoring
overhead.
The measurements are aggregated across data-parallel workers and
recorded as a single layer-by-metric table, together with global minimum
and maximum values across all MoE layers for time-series alerts.

We track the minimum, median, mean, and maximum token counts across
experts, and summarize routing balance using

\begin{equation}
    R_{\mathrm{dispatch}}
    =
    \frac{
        \min_i N_i
    }{
        \operatorname{median}_i N_i
    },
    \label{eq:dispatch_min_median}
\end{equation}

where $N_i$ is the number of tokens dispatched to expert $i$ during the
step.
Unlike a maximum or variance statistic, this ratio directly measures
how the least-used expert compares with a typical expert and is robust
to a single unusually popular expert.
Values above $0.7$ generally indicate healthy balance, whereas values
below $0.3$ indicate likely starvation.
The global minimum and maximum expert-token counts provide complementary
signals for starvation and overload.

Dispatch balance alone is insufficient to determine whether an expert
is functionally healthy.
An expert can continue receiving tokens while its contribution to the
residual stream vanishes.
We therefore monitor the norm of each expert's output projection
$\mathbf{W}_{2}^{(i)}$ and compute its minimum-to-median ratio.
The output projection is particularly informative because it is applied
after PolyNorm and directly controls the scale written back to the
residual stream.
A ratio above $0.8$ is typically healthy, while a sharp decline below
$0.5$ signals that at least one expert may be dying.

We also separate the routed and shared contributions.
Because the implementation exposes their combined output, the routed
component is recovered as

\begin{equation}
    \mathbf{y}_{\mathrm{routed}}
    =
    \mathbf{y}_{\mathrm{final}}
    -
    \mathbf{y}_{\mathrm{shared}}.
    \label{eq:routed_output_recovery}
\end{equation}

For both streams, we log the RMS, mean absolute value, and absolute
maximum, as well as the corresponding routed-to-shared ratios.
In particular,

\begin{equation}
    R_{\mathrm{routed/shared}}
    =
    \frac{
        \operatorname{RMS}(\mathbf{y}_{\mathrm{routed}})
    }{
        \operatorname{RMS}(\mathbf{y}_{\mathrm{shared}})
    }.
    \label{eq:routed_shared_rms}
\end{equation}

This ratio uses the routed-only output rather than the combined output,
which avoids an artificial additive offset and preserves cancellation
effects between the two streams.
A value near zero indicates that the shared experts perform nearly all
of the work, while values above approximately $0.3$ indicate a
meaningful routed-expert contribution.
Sudden changes in either stream's RMS indicate scale instability, and
an absolute maximum more than roughly ten times its RMS provides an
early warning of activation outliers.

Finally, we measure functional expert diversity using output
activations rather than parameter vectors.
For each expert, we average its output over the tokens routed to it and
compute all pairwise cosine similarities between the resulting mean
vectors.
We log both the full similarity distribution and its maximum.
This activation-based statistic directly detects experts that behave
similarly on their assigned tokens, whereas cosine similarity between
high-dimensional flattened weight matrices is generally uninformative.
A maximum similarity below $0.5$ is typically healthy, while a value
above $0.8$ indicates severe functional redundancy.
Table~\ref{tab:moe_health_metrics} summarizes the failure modes
indicated by these expert-health metrics.

\begin{table}[t]
    \centering
    \small
    \caption{
        Principal expert-health signals monitored during Motif 3
        pretraining.
    }
    \label{tab:moe_health_metrics}
    \begin{tabular}{p{0.29\linewidth}p{0.61\linewidth}}
        \toprule
        \textbf{Metric} & \textbf{Failure mode indicated} \\
        \midrule
        Dispatch min/median
        &
        Starvation or a dead expert caused by routing imbalance
        \\
        Maximum expert-token count
        &
        Concentrated traffic and possible expert overload
        \\
        Output-weight min/median
        &
        Hidden collapse in which an expert receives tokens but its
        output projection contributes increasingly little
        \\
        Routed/shared RMS, mean, and maximum
        &
        Dominance of shared experts, routed-expert collapse, or a
        change in the routed-output distribution
        \\
        Maximum expert-output cosine similarity
        &
        Functional collapse in which nominally distinct experts learn
        similar representations
        \\
        Routed and shared abs-max/RMS
        &
        Activation outliers and early warning signs of numerical
        instability
        \\
        \bottomrule
    \end{tabular}
\end{table}

Persistent imbalance triggers the selective recovery procedure
described above, including immediate replacement from an adjacent
stable router when a routing collapse is detected.
By jointly monitoring dispatch, parameter, and activation statistics,
we can detect both visible routing collapse and hidden failure modes in
which token counts remain balanced but expert specialization has
already deteriorated.

\subsection{Training Setups}
\label{sec:pretraining_setups}

We train Motif 3 using the standard autoregressive next-token
prediction objective. Its sparse MoE architecture provides
approximately 314 billion parameters of total capacity while activating
approximately 13.2 billion parameters for each token. This sparsity
allows the model to maintain a large pool of experts without incurring
the per-token computation of a comparably sized dense model. We combine
the language-modeling objective with the routing-balancing and
numerical-stabilization mechanisms described in
Section~\ref{sec:moe_stabilization} to train this large expert pool
reliably and efficiently.

\paragraph{Model configuration.}
The pretraining model contains 53 Transformer layers with a model
dimension of 4,096. The first two layers use dense feed-forward
networks with an intermediate dimension of 12,288, while the remaining
51 layers use sparse MoE feed-forward networks. Each MoE layer contains
384 routed experts and one shared expert. The router uses normalized
sigmoid scores and selects eight routed experts per token. Each routed
expert has an intermediate dimension of 1,280.

GDLA uses 80 query heads and 16 KV heads, with a query-key head
dimension of 192, a value-head dimension of 128, and a 64-dimensional
rotary component. The query and KV low-rank dimensions are 1,024 and
512, respectively. We use a grouped ratio of $g=4$, yielding 64 signal
and 16 noise query heads. Both query paths use the same set of 16 KV
heads. Every fourth layer uses full causal attention, and
the remaining layers use sliding-window attention with a window of 128
tokens. Under zero-based indexing, layers $0,4,8,\ldots$ use full
attention, while the three intervening layers use sliding-window
attention.

\paragraph{Batch size and context-length schedule.}
The global training batch contains up to 75 million tokens. Pretraining
begins with sequences of up to 4K tokens. During the learning-rate
decay phase, we transition to a maximum sequence length of 32K tokens,
followed by a dedicated long-context stage with sequences of up to
256K tokens. The full-attention layers use a DeepSeek-YaRN extension
from the original 4K context length to 256K, with an extension factor
of 64, while the sliding-window layers retain local RoPE without
long-range extension.

For the long-context stage, we reconstruct approximately 5\% of the
overall pretraining corpus as a dedicated long-context dataset. We
limit reasoning-focused data to less than 5\% of the pretraining
mixture so that the model develops reasoning capability without
overconcentrating the base-model distribution on reasoning traces.

\paragraph{Hyperparameters.}
Training starts with a learning rate of $3\times10^{-4}$ and follows a
warmup-stable-decay (WSD) learning-rate schedule. The stable phase
performs the majority of pretraining at the maximum learning rate,
after which the decay phase reduces the learning rate to
$7.5\times10^{-5}$ while transitioning from 4K to 32K sequences. The
subsequent 256K long-context stage continues from this checkpoint and
further reduces the learning rate to $3\times10^{-5}$.

The MTP auxiliary objective uses a loss weight of $0.2$. The
auxiliary-loss-free expert-selection bias uses a balancing coefficient
of $1\times10^{-3}$. The sequence-wise auxiliary load-balancing loss is
summed over tokens and uses a default weight of $1\times10^{-4}$. We
increase this weight to $2.4\times10^{-4}$ for layer 2 and to
$2.0\times10^{-4}$ for layers 3--5. The FFN magnitude regularizer uses
a loss weight of $2.0\times10^{-4}$ and a threshold of $128$, while the
final layer uses a threshold of $1{,}024$ with the same loss weight.

\subsection{Pretraining Evaluation}
\label{sec:pretraining_evaluation}

We evaluate the final pretrained base model on a compact set of
knowledge, commonsense reasoning, mathematics, and code-generation
benchmarks. Table~\ref{tab:pretraining_evaluation} reports the absolute
performance of the checkpoint under the indicated prompting settings.
We present these results primarily to characterize the capabilities
acquired during pretraining, without attempting a direct comparison
across models evaluated with potentially different harnesses and
prompting protocols.

The evaluation covers knowledge and science with
MMLU~\cite{hendrycks2021mmlu},
MMLU-Pro~\cite{wang2024mmlupro}, and ARC-C~\cite{clark2018arc};
commonsense reasoning with WinoGrande~\cite{sakaguchi2020winogrande},
HellaSwag~\cite{zellers2019hellaswag}, and PIQA~\cite{bisk2020piqa};
mathematics with GSM8K~\cite{cobbe2021gsm8k} and
MATH~\cite{hendrycks2021math}; and code generation with
HumanEval~\cite{chen2021humaneval} and MBPP~\cite{austin2021mbpp}.

We report accuracy for the multiple-choice and mathematics benchmarks
and pass@1 for HumanEval and MBPP.

\begin{table}[t]
    \centering
    \small
    \caption{Evaluation results for the Motif 3 pretrained base model.
    CoT denotes chain-of-thought prompting.}
    \label{tab:pretraining_evaluation}
    \resizebox{\linewidth}{!}{%
    \begin{tabular}{cccccccccc}
        \toprule
        \makecell{\textbf{MMLU} \\ 5-shot}
        & \makecell{\textbf{MMLU-Pro} \\ 5-shot CoT}
        & \makecell{\textbf{ARC-C} \\ 25-shot}
        & \makecell{\textbf{WinoGrande} \\ 5-shot}
        & \makecell{\textbf{HellaSwag} \\ 10-shot}
        & \makecell{\textbf{PIQA} \\ 0-shot}
        & \makecell{\textbf{GSM8K} \\ 8-shot CoT}
        & \makecell{\textbf{MATH} \\ 4-shot CoT}
        & \makecell{\textbf{HumanEval} \\ 0-shot}
        & \makecell{\textbf{MBPP} \\ 3-shot} \\
        \midrule
        86.20
        & 68.56
        & 94.71
        & 80.90
        & 88.30
        & 85.14
        & 93.93
        & 70.58
        & 73.70
        & 84.60 \\
        \bottomrule
    \end{tabular}
    }
\end{table}
\section{Post-training}

Our post-training pipeline consists of three stages: general supervised
fine-tuning (SFT), specialized teacher training, and Multi-teacher
On-Policy Distillation (MOPD).

We first train a preliminary SFT model to identify capability-specific
failure modes and guide targeted data generation. This preliminary model
is used only for data construction and is not used as the initialization
of the final student model. After constructing the complete SFT corpus,
we restart from the pretrained checkpoint and train the general SFT
student on the full data mixture.

We then derive seven specialized teacher models from the general SFT
checkpoint: six through domain-specific reinforcement learning and one
software-engineering teacher through SFT. Finally, MOPD
transfers the complementary capabilities of these teachers back into the
general SFT student, producing a unified model that retains broad general
capabilities while acquiring targeted expertise.

\subsection{Supervised Fine-Tuning}
\label{sec:posttraining_sft}

\subsubsection{Data}

We construct the SFT corpus primarily from selected datasets in the
Nemotron family~\cite{nvidia2026nemotronposttrainingv3,
ahmad2026openswetraces,pi2026dataengineeringscalingllm}, supplemented
with broadly collected and synthetically generated data. The resulting
mixture covers instruction following, reasoning, coding, knowledge, and
agentic tasks. For tool-use examples, we retain the interaction context
across successive tool calls so that the training sequences represent
complete problem-solving trajectories.

For agentic data, a preliminary SFT model identifies failure-prone
decisions in unsuccessful trajectories. We generate related instances
and obtain corrected actions or continuations conditioned on the
preceding interaction history. These examples are combined with complete
agent trajectories and the rest of the corpus. The preliminary model is
used only for data construction; the final student is trained from the
pretrained checkpoint on the consolidated corpus.

\subsubsection{Training Setup}

We train the general SFT student from the pretrained checkpoint,
following the optimization setup used during pretraining. Training is
performed in stages with different data mixtures and uses a maximum
sequence length of 256K tokens.

We use a global batch size of 33 million tokens and initially maintain the learning
rate at $3 \times 10^{-5}$. During the final phase of training, we apply
cosine decay to reduce the learning rate to $1.5 \times 10^{-6}$.

We retain the pretraining stabilization settings, except that the
expert-selection bias coefficient and sequence-wise auxiliary
load-balancing loss weights are each reduced by a factor of ten. The
expert-selection bias coefficient is therefore $1\times10^{-4}$. The
sequence-wise auxiliary loss uses a default weight of $1\times10^{-5}$,
with weights of $2.4\times10^{-5}$ for layer 2 and $2.0\times10^{-5}$
for layers 3--5. The MTP objective uses a loss weight of $0.1$.
The FFN magnitude regularizer retains a loss weight of
$2.0\times10^{-4}$ and a threshold of $128$, with a threshold of
$1{,}024$ for the final layer.

As in pretraining, we retain the shared-weight multi-token prediction
(MTP) objective during SFT. We apply assistant-only loss masking to both
the standard next-token prediction objective and the MTP objective.
Specifically, each prediction contributes to the loss only when its
corresponding target token belongs to an assistant response.

\subsection{Reinforcement Learning}

\subsubsection{TorchTitan Backend for NeMo RL}
\label{sec:rl_torchtitan_backend}

Motif 3 is pretrained and supervised fine-tuned using our in-house
TorchTitan training stack~\cite{liang2025torchtitan}. For reinforcement learning, we build on
NVIDIA NeMo RL, an open-source post-training library designed for
efficient reinforcement learning at scale
~\cite{nvidianemorl}. Its native training backends are Megatron Core and
AutoModel (DTensor). The verifiers and interactive environments used for
training and evaluation are provided separately through NVIDIA NeMo Gym,
an open-source library for evaluating and improving models and agents
through environments~\cite{nvidianemogym}. Reimplementing Motif 3 in a
native NeMo RL backend would introduce a second model implementation and
create a risk of divergence from the model used during pretraining and
SFT.

We instead implement TorchTitan as an external NeMo RL training backend.
The integration attaches the required trainer and model interfaces at
import time, without modifying the upstream NeMo RL codebase. The same
Motif 3 implementation and checkpoints used for pretraining and SFT can
therefore be optimized directly with NeMo RL's GRPO algorithm and
evaluated by NeMo Gym verifiers.

The backend supports the multi-dimensional parallelism used by our
TorchTitan stack, including tensor, context, pipeline, and data
parallelism with FSDP2. It supports both colocated and non-colocated
rollout generation with vLLM~\cite{kwon2023vllm} or SGLang~\cite{zheng2024sglang}. We perform rollout inference
with block-wise FP8 quantization, reducing the memory footprint of the
rollout workers and improving generation throughput. Multi-node training
and rollout orchestration are managed through Ray~\cite{moritz2018ray}.

\subsubsection{Asynchronous GRPO and In-Flight Weight Updates}
\label{sec:async_grpo}

For the specialist RL runs, generation and training execute on disjoint
GPU pools as separate Ray actors. An asynchronous collector continuously
issues vLLM rollouts while the learner trains on trajectories drawn from
a replay buffer. Each trajectory is tagged with the policy weight
version that generated it and the learner step for which it is eligible.
We bound trajectory age to one optimizer step, discard samples that
exceed this bound, and wait when the target-version batch is not yet
available. This design overlaps rollout generation with policy
optimization while preventing unbounded policy staleness.

After each optimizer step, the updated policy weights are broadcast to
the vLLM workers without first draining all pending rollout requests.
Before the distributed weight-update collective, the generation workers
briefly pause at a common wave boundary. In-flight requests and their KV
caches remain resident during the pause, avoiding collective-ordering
deadlocks while allowing generation to resume immediately after the
refit. In our runs, we retain the existing KV cache rather than
recomputing it after each update.

Because asynchronously generated trajectories can differ from the
current learner policy, we cache their behavior-policy log probabilities
and apply token-level importance-sampling correction. The importance
weights are truncated to $[0.2,5.0]$ to limit the variance caused by
policy staleness. Together, the one-step age bound, in-flight refitting,
and truncated importance sampling provide generation-training overlap
while keeping updates close to the current policy.

\subsubsection{Specialized Teacher Training}
\label{sec:specialized_teacher_training}

Starting from the same general SFT checkpoint, we train six specialist
teachers with GRPO and one software-engineering teacher with SFT. The RL
tasks span 13 verifier domains. We do not optimize a single policy
against every verifier simultaneously,
because verifier latency and reward variance differ substantially across
domains, and mixing unrelated reward surfaces couples otherwise distinct
failure modes. Instead, we group the 13 domains into six GRPO jobs. A
teacher covers one domain when that domain is sufficiently large to
support a dedicated run, and combines several domains when their tasks
and reward structures are closely related. Table~\ref{tab:rl_specialist_teachers}
summarizes the seven teachers and their capability coverage.

\begin{table}[t]
    \centering
    \small
    \caption{The seven specialist teachers and their capability coverage.}
    \label{tab:rl_specialist_teachers}
    \begin{tabular}{p{0.25\linewidth}p{0.65\linewidth}}
        \toprule
        \textbf{Teacher} & \textbf{Coverage} \\
        \midrule
        Agentic tool use
        & Interactive shell and tool environments; multi-step task execution \\
        Professional work
        & Open-ended occupational deliverables graded by comparison \\
        Software engineering
        & Repository-level modifications verified through test execution \\
        Long-context reasoning \& abstention
        & Retrieval and synthesis over very long inputs; calibrated abstention \\
        Mathematics
        & Competition and proof-style problems; symbolic and judged answer checking \\
        Code and science
        & Program synthesis, scientific computing, and physical reasoning \\
        Chat
        & Dialogue quality, instruction following, and safety \\
        \bottomrule
\end{tabular}
\end{table}

\paragraph{General optimization recipe.}
Most single-turn RL teachers use a token-level GRPO objective
based on DAPO~\citep{yu2025dapo}. Depending on the cost and structure of each domain, each
optimization step uses a global batch ranging from 256 to 1,152
trajectories, with the numbers of prompts and generations per prompt
adjusted accordingly. Rewards are normalized within each group, and
advantages are computed with a leave-one-out baseline and clipped to the
interval $[-4,4]$. All specialist-teacher runs use a learning rate of
$3\times10^{-6}$. The maximum rollout length ranges from 8K to 192K
tokens across domains. The exact episode budget and number of steps vary
across specialist domains.

To improve learner efficiency, we apply sequence packing to the
collected rollout trajectories. We use a modified first-fit decreasing
algorithm to pack shorter trajectories into token-balanced microbatches,
with maximum sequence lengths of up to 192K tokens depending on the
domain. Sequence packing is applied to both the training forward pass
and behavior-policy log-probability computation. This reduces padding
and shortens the forward-pass time for batches containing trajectories
of different lengths.

We use DAPO-style asymmetric ratio clipping with a lower clipping width
of $0.2$ and an upper clipping width of $0.28$, corresponding to a
policy-ratio interval of $[0.8,1.28]$. The loss is computed at the token
level. Overlong responses are excluded from the loss. We do not use
entropy regularization, and use a small reference-policy KL term where an
additional anchor to the SFT policy is required.

\paragraph{Frozen MTP and routing parameters.}
Throughout both specialist-teacher training and MOPD, we freeze the
shared-weight MTP parameters, the MoE router weights, and the
expert-selection biases. This preserves the routing configuration and
the auxiliary prediction head learned during SFT while the expert and
dense model parameters are optimized. Although the policy continues to
change around the fixed MTP module, we observe no measurable degradation
in the MTP draft-token acceptance rate over the course of training.
Because the MTP parameters remain frozen during distillation, the final
MOPD model retains exactly the same MTP weights as the student used to
initialize MOPD.

\paragraph{Offline prompt filtering.}
We disable online dynamic sampling. Instead, before RL teacher training, we
perform four to eight rollout attempts per candidate prompt, depending
on the dataset, and estimate its empirical pass rate $p$. We retain
prompts satisfying $0 < p \leq 0.8$, removing both prompts that are never
solved by the initial policy and prompts that are already solved too
consistently. Training then uses this fixed filtered pool. This
front-loaded filtering concentrates optimization on nontrivial but
learnable examples without coupling the training loop to an online
dynamic-sampling procedure.

\paragraph{Agentic tool-use teacher.}
We train the agentic tool-use teacher on end-to-end terminal tasks. Each
task is instantiated in an isolated, task-specific container environment
built from its own Dockerfile. The model interacts with the terminal
until it produces a final result or exhausts its episode budget of at
most 100 interaction turns and one hour of wall-clock time. Rewards are
provided by task-specific outcome verifiers, with no intermediate
process supervision: a successful trajectory receives 1 and an
unsuccessful trajectory receives 0.

\paragraph{Professional-work teacher.}
We train the professional-work teacher on long-horizon workplace tasks
that require producing spreadsheets, slide decks, and written reports.
Each task is instantiated in an isolated container environment equipped
with tools for document creation and editing. The model gathers
information and creates and revises files until it produces its final
documents or exhausts its episode budget of at most 100 interaction
turns and 45 minutes of wall-clock time.

Because document quality cannot be assessed reliably with deterministic
verifiers, rewards are preference-based. An LLM judge compares each
generated document with a reference according to task-specific quality
criteria, and we perform the comparison in both candidate orders to
reduce position bias. A trajectory receives 1.0 when it wins more of
these judgments than the reference does, 0.0 when the reference wins
more, and 0.5 otherwise, with ties counting toward neither side. We
select the teacher checkpoint using periodic held-out evaluation.

\paragraph{Software-engineering teacher.}
The software-engineering teacher is trained with SFT. Starting from the general SFT model, we continue fine-tuning on repository-level software-engineering tasks with a learning rate of $5 \times 10^{-6}$, while keeping the router parameters frozen throughout training. We construct training tasks from pull requests in public Git repositories and exclude tasks whose base commits overlap with those in SWE-bench. For supervision, we retain only successful model-generated trajectories whose repository modifications pass the corresponding tests. We further filter out trajectories that retrieve or reproduce the gold patch from Git repositories or other external sources, ensuring that the retained solutions are independently produced by the model.

\paragraph{Long-context reasoning and abstention teacher.}
This teacher combines retrieval and synthesis over very long inputs with
calibrated abstention. We construct the training data from difficult,
open-ended questions that even strong frontier systems, including GPT
and Claude Opus models, frequently fail to answer reliably. These tasks
train the model to distinguish between questions supported by sufficient
evidence and questions for which abstention is preferable to speculation.

We use a deterministic LLM judge that assigns one of four grades:
correct, partial answer, not attempted, or incorrect. The
corresponding rewards are $1$, $0.666$, $0.333$, and $0$, respectively.
This graded ladder prioritizes correctness while rewarding an explicit
abstention over an incorrect, potentially hallucinated answer.

Incomplete or empty generations receive zero reward without invoking the
judge. This prevents the policy from exploiting the abstention reward by
failing to finish an answer. The judge is required to return a single
grade, and malformed or unparseable outputs are conservatively treated
as incorrect.

\paragraph{Mathematics teacher.}
This teacher is trained to solve mathematical problems ranging from
competition-level questions to research-level problems. We verify
answers with symbolic and structured mathematical checking libraries,
including \texttt{math-verify},\footnote{\url{https://github.com/huggingface/Math-Verify}} whenever their output formats permit
reliable automatic comparison. For problems that cannot be validated by
these libraries, such as open-ended proofs or answers requiring
mathematical judgment, we use an LLM judge to determine correctness.

\paragraph{Code-and-science teacher.}
This teacher uses coding tasks ranging from competitive programming to
research-level scientific computing. Its training mixture also includes
scientific knowledge, numerical and symbolic scientific computation,
and physical reasoning across a broad range of difficulty levels. The
scientific-computing data range from subproblems that must be solved as
part of a larger main problem to self-contained standalone problems. We
generate task-specific test cases for these problems and verify model
outputs through direct execution. We also use domain-specific scientific
libraries whenever possible. For tasks that cannot be reliably assessed
by these automatic verifiers, we use an LLM judge to evaluate
correctness.

\paragraph{Chat teacher.}
This teacher targets dialogue quality, instruction following, and
safety. Its prompt mixture combines curated and synthetically generated
prompts. A substantial portion of the training data focuses on safe and
appropriate responses to harmful, sensitive, or otherwise unsafe
requests while preserving helpfulness on benign queries. The mixture
also places substantial emphasis on Korean responses. For the Korean
chat tasks, we use a two-stage reward.
We first apply a deterministic LLM judge with a strict Korean-language
rubric. A response receives a base reward of 1 only when it passes every
rubric criterion; violation of any single criterion results in a reward
of 0. Empty responses and unparseable or ambiguous judge verdicts also
receive 0.

The default rubric rejects eight anti-patterns: inappropriate English
code switching or romanization, foreign-script mixing, a mismatch with
the requested language, repetition or degeneration, failure to answer,
off-topic content, low-effort or excessively short responses, and
non-polite Korean. The code-switching criterion is directional: a
Korean term followed by an English gloss is permitted, whereas an
English term followed by a Korean gloss is treated as unnecessary
mixing. One training variant omits only the politeness criterion and
retains the other seven anti-patterns.

Responses with a base reward of 0 remain at 0. For a response that
passes the full rubric, we apply a piecewise-constant shaping function
based on its token length $L$:
\begin{equation}
r_{\mathrm{chat}}(L)=
\begin{cases}
0.5, & L < 256, \\
1.0, & 256 \leq L < 512, \\
1.5, & 512 \leq L < 1024, \\
1.0, & L \geq 1024.
\end{cases}
\end{equation}
The peak followed by a lower reward for very long answers is
intentional. Under within-group reward normalization, a rank-order
change is more effective at controlling response length than a small
continuous penalty. Although this training is primarily Korean-centered, we
observe improved instruction-following performance in both Korean and
English.

\begin{figure*}[t]
    \centering
    \includegraphics[width=\textwidth]{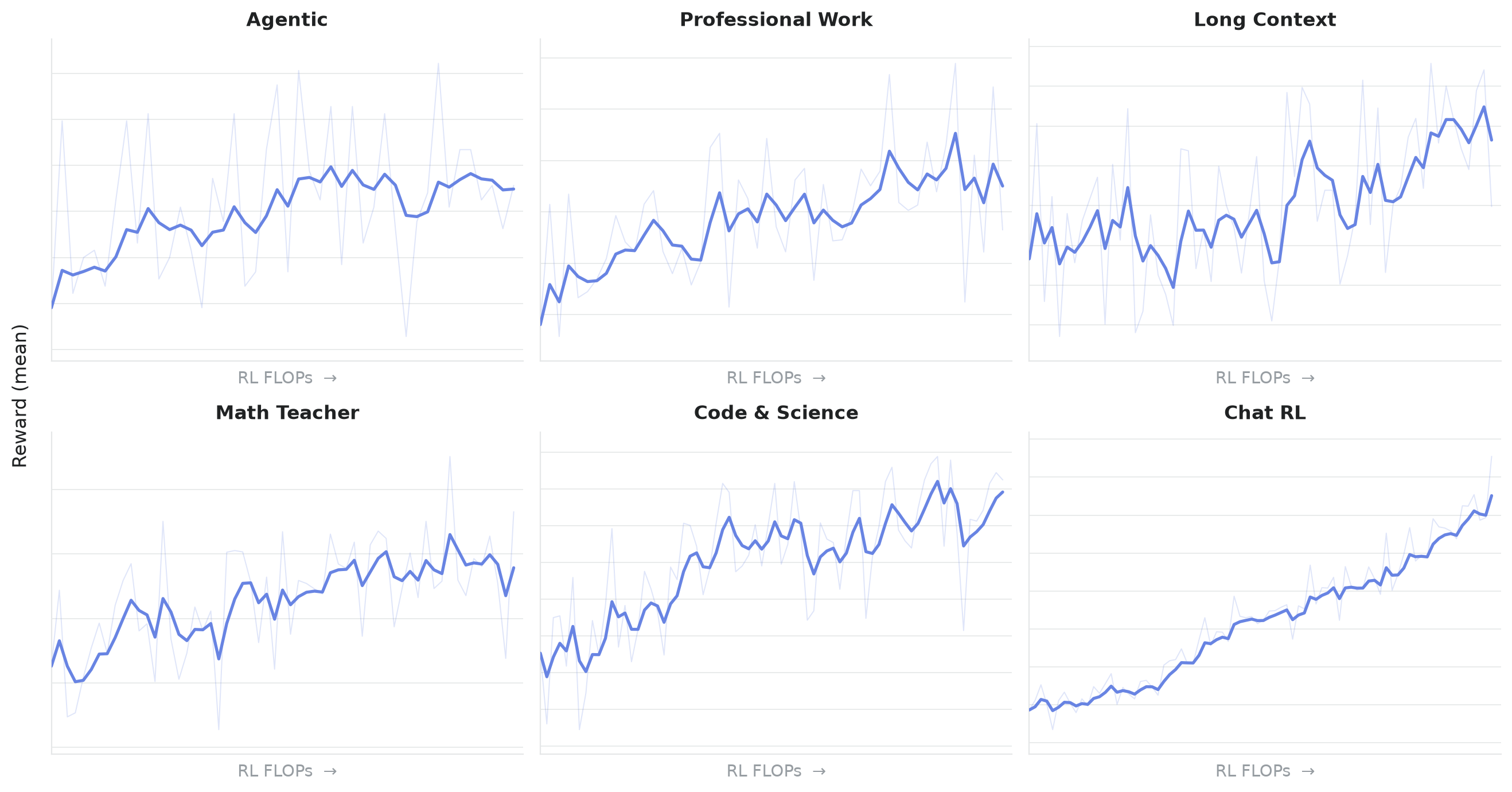}
    \caption{Mean reward over cumulative RL compute for the six
    GRPO-trained specialist teachers. The lighter curves show the per-update reward
    measurements, and the darker curves show their smoothed trends.}
    \label{fig:specialist_teacher_rewards}
\end{figure*}

Figure~\ref{fig:specialist_teacher_rewards} shows the reward trajectories
for the six GRPO-trained specialist teachers over cumulative RL compute.

\subsubsection{Multi-teacher On-Policy Distillation}
\label{sec:mopd}

After specialized teacher training, we apply Multi-teacher On-Policy
Distillation to consolidate their capabilities into a single model. The
general SFT model initializes the student, and the seven specialists
provide supervision for their respective target domains. We generate
on-policy trajectories with the student, route each example to its
domain specialist, and disable environment grading so that verifier
rewards do not enter the optimization objective.

Our implementation can distill full-vocabulary probability
distributions, but we do not use them in this work. For each token
sampled by the student, we evaluate only the scalar log probability
assigned to that token by the routed teacher. We merge these teacher log
probabilities row-wise so that each trajectory receives supervision from
its corresponding specialist.

To control mismatch between the generation and training policies, we
first compute the token-level importance weight
\begin{equation}
    \widetilde{w}_t =
    \begin{cases}
        w_t, & 0.5 \leq w_t \leq 5.0, \\
        0, & \text{otherwise},
    \end{cases}
    \qquad
    w_t = \frac{\pi_{\mathrm{old}}(y_t\mid x,y_{<t})}
    {\pi_{\mathrm{gen}}(y_t\mid x,y_{<t})}.
    \label{eq:mopd_icepop}
\end{equation}
This ICE-POP filter~\cite{lingteam2025everystep} drops tokens whose importance weights fall outside
the specified interval. The detached OPD signal is
\begin{equation}
    d_t = \operatorname{sg}\!\left[
        \log \frac{\pi_{T(x)}(y_t\mid x,y_{<t})}
        {\pi_{\mathrm{old}}(y_t\mid x,y_{<t})}
    \right].
    \label{eq:mopd_logprob_ratio}
\end{equation}
The resulting token-level distillation loss is
\begin{equation}
    \mathcal{L}_{\mathrm{MOPD}} =
    -\mathbb{E}_t\!\left[
        \widetilde{w}_t\,d_t\,
        \log\pi_\theta(y_t\mid x,y_{<t})
    \right],
    \label{eq:mopd_opd_loss}
\end{equation}
where $T(x)$ selects the teacher for example $x$ and
$\pi_{\mathrm{old}}$ is the student before the update. Environment
rewards and the reference-policy KL term are both omitted.

We use a global batch of 512 on-policy trajectories, with one generation
per prompt. We use the Muon optimizer with no weight decay and decay the
learning rate from $5\times10^{-6}$ to $3\times10^{-6}$ over the course
of distillation. The maximum sequence length is 163,840 tokens.

\subsection{Evaluation}

\subsubsection{Evaluation Setup}
For contextual comparison, we compare Motif 3 with strong open-weight models using scores reported on the corresponding benchmark leaderboards. All evaluations for Motif 3
were performed with a sampling temperature $= 1.0$, $\text{top-p} = 0.95$ and a maximum sequence length of 262,144 tokens.
Benchmark-specific evaluation settings are provided in Appendix~\ref{sec:evaluation-details}.

\paragraph{Agentic}
We evaluate agentic capabilities using GDPval-AA v2~\cite{patwardhan2026gdpval,artificialanalysis2026gdpvalaa},
$\tau^2$-Bench Telecom~\cite{barres2025tau}, $\tau^3$-Banking~\cite{shi2026tauknowledge,artificialanalysis2026tau3banking}, and ITBench-AA~\cite{jha2025itbench,artificialanalysis2026itbenchaa}.
These benchmarks assess economically valuable professional task
completion, user--agent coordination in dual-control telecom support,
knowledge-grounded tool use in banking workflows, and root-cause
diagnosis of Kubernetes incidents, respectively.

\paragraph{Coding}
We assess coding, software-engineering, and terminal-based
problem-solving capabilities using SWE-bench Verified~\cite{jimenez2024swe}, Terminal-Bench
2.1~\cite{merrill2026terminal}, and SciCode~\cite{tian2024scicode}. These benchmarks evaluate the ability to resolve
real-world repository issues, complete diverse tasks in terminal
environments, and generate code for scientific research problems,
respectively.

\paragraph{Reasoning and Knowledge}
We evaluate mathematical and scientific reasoning, expert-level
knowledge, and factual reliability using IMO-AnswerBench~\cite{luong2025towards}, Apex
Shortlist~\cite{dekoninck2026beyond}, GPQA Diamond~\cite{rein2023gpqa}, HLE~\cite{phan2025humanity}, CritPt~\cite{zhu2025probing}, and AA-Omniscience~\cite{jackson2025aa}. Together, these
benchmarks cover olympiad-level mathematics, graduate-level science,
broad expert knowledge, research-level physics reasoning, and factual
recall and calibration. For AA-Omniscience, we report both accuracy and
the non-hallucination score used in our evaluation.

\paragraph{Long Context and Instruction Following}
We evaluate long-context reasoning using AA-LCR~\cite{artificialanalysis2025lcr} and precise instruction
following using IFBench~\cite{pyatkin2026generalizing}. AA-LCR requires models to integrate and reason
over information distributed across multiple long documents, whereas
IFBench evaluates generalization to diverse, verifiable output
constraints.

\newcommand{\NA}{--}
\newcommand{\subbench}[1]{\hspace{1.2em}#1}

\newcommand{\sectionrow}[1]{%
  \specialrule{0.6pt}{2pt}{0pt}%
  \rowcolor{gray!20}[0pt][0pt]%
  \multicolumn{7}{@{}l@{}}{\hspace{2pt}\bfseries\strut #1}\\%
  \specialrule{0.4pt}{0pt}{2pt}%
}

\begin{table*}[tb]
\centering

\scriptsize
\setlength{\tabcolsep}{4.2pt}
\renewcommand{\arraystretch}{1.05}

\resizebox{\linewidth}{!}{%
\begin{tabular}{
  @{}l
  @{\hspace{5pt}\vrule width 0.4pt\hspace{5pt}}
  *{6}{c}
  @{}
}
\toprule

\makecell[c]{\textbf{Benchmark}}
& \makecell[c]{\textbf{Motif 3}\\\textbf{314B-A13B}}
& \makecell[c]{\textbf{MiniMax-3}\\\textbf{428B-A23B}}
& \makecell[c]{\textbf{GLM-5.1}\\\textbf{744B-A40B}}
& \makecell[c]{\textbf{Kimi-K2.6}\\\textbf{1T-A32B}}
& \makecell[c]{\textbf{Qwen-3.7}\\\textbf{Max}}
& \makecell[c]{\textbf{DS-v4-Pro}\\\textbf{1.6T-A49B}}
\\

\midrule

\sectionrow{Agentic}

GDPval-AA v2
& 38.7 & 44.4 & 37.8 & 34.4 & 39.0 & 40.2 \\

$\tau^2$-Bench Telecom
& 94.7 & 88.9 & 97.7 & 95.9 & 94.7 & 96.2 \\

$\tau^3$-Banking
& 35.3 & 15.3 & 13.6 & 23.3 & 12.0 & 30.1 \\

ITBench-AA
& $51.5^*$ & - & 40.3 & 31.2 & 42.5 & 38.3 \\

\sectionrow{Coding}

SWE-bench Verified
& 76.2 & 75.0 & 76.4 & 76.2 & 80.4 & 77.4 \\

Terminal-Bench 2.1
& 74.9 & 65.2 & 61.8 & 65.9 & 75.0 & 64.0 \\

SciCode
& 40.6 & 45.4 & 43.8 & 53.5 & 53.5 & 50.0 \\

\sectionrow{Reasoning and Knowledge}

IMO-AnswerBench
& 83.2 & - & 83.8 & 81.8 & 90.0 & 89.8 \\

Apex Shortlist
& 75.5 & - & 71.1 & 77.4 & 44.5 & 85.8 \\

GPQA Diamond
& 83.4 & 92.9 & 86.8 & 91.1 & 92.4 & 88.8 \\

HLE
& 37.0 & 39.0 & 30.1 & 37.5 & 41.4 & 37.5 \\

CritPt
& 6.6 & 3.7 & 4.6 & 8.0 & 11.4 & 12.9 \\

AA-Omniscience Accuracy
& 30.1 & 16.7 & 23.7 & 32.6 & 31.0 & 42.9 \\

AA-Omniscience Non-Hallucination
& 71.6 & 81.6 & 70.1 & 59.5 & 74.0 & 5.9 \\

\sectionrow{Long Context and Instruction Following}

AA-LCR
& 72.3 & 80.3 & 68.0 & 76.7 & 75.0 & 70.0 \\

IFBench
& 78.2 & 82.9 & 76.3 & 76.0 & 79.1 & 76.5 \\

\bottomrule
\end{tabular}%
}

\caption{
Evaluation results for Motif 3. An asterisk (*) indicates that the corresponding result for Motif 3 is evaluated on the public subset only.
}
\label{tab:evaluation-suite}

\end{table*}

\subsubsection{Results}

As shown in Table~\ref{tab:evaluation-suite}, Motif 3 performs particularly well on agentic and tool-oriented benchmarks. It achieves the highest score among the models and results listed in Table~\ref{tab:evaluation-suite} on $\tau^3$-Banking (35.3), a near-best score of 74.9 on Terminal-Bench 2.1, and 94.7 on $\tau^2$-Bench Telecom. On broader professional and IT-oriented evaluations, Motif 3 scores 38.7 on GDPval-AA v2 and 51.5 on the public subset of ITBench-AA. The latter is the highest among the available results listed in Table~\ref{tab:evaluation-suite}.
These results indicate that Motif 3 is particularly effective at multi-step task execution and interaction with external tools and environments.

Motif 3 also maintains competitive performance across coding, mathematical reasoning, and general knowledge benchmarks. It achieves 76.2 on SWE-bench Verified and 40.6 on SciCode.
On reasoning-oriented evaluations, Motif 3 scores 83.2 on IMO-AnswerBench, 75.5 on Apex Shortlist, 83.4 on GPQA Diamond, and 37.0 on HLE. Although these results are generally comparable to those of other models in the comparison, performance on SciCode and CritPt remains below the strongest models, suggesting further room for improvement in scientific coding and specialized scientific reasoning.

On AA-Omniscience, Motif 3 achieves an accuracy score of 30.1 and a non-hallucination score of 71.6.
Its accuracy remains below the best-performing models, whereas its non-hallucination score is among the highest results listed in Table~\ref{tab:evaluation-suite}. This result suggests a relatively favorable balance between answering scientific questions correctly and avoiding unsupported answers. Motif 3 further achieves 72.3 on AA-LCR and 78.2 on IFBench, showing competitive performance on long-context reasoning and instruction following.

Overall, Motif 3 exhibits broad performance across the evaluation suite, with its clearest strengths in agentic execution and terminal-based problem solving. These results are consistent with the objective of MOPD, which consolidates specialized teachers spanning agentic tool use, professional work, software engineering, long-context reasoning and abstention, mathematics, code and science, and chat into a single student model.
Motif 3 consequently maintains competitive performance across diverse capabilities while achieving particularly strong results on agentic and tool-oriented tasks.

\section{Conclusion, Limitations, and Future Directions}

We presented Motif 3, a 314-billion-parameter Mixture-of-Experts
language model that activates approximately 13.2 billion parameters per
token. Its fine-grained MoE design provides 384 routed experts per
sparse layer while activating only eight for each token, combining a
large expert pool with limited per-token expert computation. Motif 3
combines this highly sparse expert architecture with GDLA, modified
manifold-constrained hyper-connections, Expert-Specific PolyNorm, and
multi-token prediction. These components are complemented by careful
optimization of routing, numerical stability, low-precision computation,
communication, memory use, and long-context parallelism. Together, the
architectural and systems design enables efficient training and
inference while retaining the capacity of a substantially larger model.

Our post-training pipeline further combines general supervised
fine-tuning, capability-specific teacher training, and Multi-teacher
On-Policy Distillation. This pipeline consolidates specialized
capabilities into a single model without requiring separate models at
deployment time. Across the reported evaluation suite, Motif 3 achieves
strong performance in agentic tool use, terminal-based problem solving,
reasoning, coding, long-context understanding, and instruction
following. These results indicate that sparse scaling, careful
stabilization, and targeted capability transfer can jointly produce a
model that is both computationally efficient and broadly capable.

The present model nevertheless has several limitations. Its training
and evaluation do not cover the full diversity of real-world tasks,
domains, languages, interaction patterns, and deployment conditions.
Performance can therefore vary on tasks that are underrepresented in
the training mixture or absent from the evaluation suite. Motif 3 is
also primarily a text model, which limits its applicability to tasks
that require direct understanding of visual inputs. Finally, although
the model supports long contexts, many long-horizon applications require
reliable state tracking, planning, recovery, and environment interaction
over substantially longer trajectories than those evaluated here.

Future work will extend Motif along several directions. We plan to
investigate new architectures that further reduce training and inference
costs and that scale reliably to models larger than Motif 3. We also aim
to extend native context lengths beyond one million tokens while
preserving both computational efficiency and effective use of distant
information. Another major direction is to add visual capabilities for
image and video inputs, enabling the model to address a wider range of
multimodal and visually grounded tasks. Finally, we plan to strengthen
long-horizon agent capabilities through richer environments, longer
interaction trajectories, improved planning and memory, and more robust
learning from execution outcomes.

\section*{Acknowledgments}

This work was supported by the Ministry of Science and ICT (MSIT),
Republic of Korea, through the National IT Industry Promotion Agency
(NIPA). This research was also conducted as part of the Sovereign AI
Foundation Model Project (Data Track), organized by MSIT and supported
by the National Information Society Agency (NIA), Republic of Korea.

\bibliographystyle{plainnat}
\bibliography{references}

\clearpage

\appendix

\section{Tokenizer}
\label{app:tokenizer}

\subsection{Pre-tokenization Regexes}
\label{app:tokenizer_regexes}

The two tokenizer training stages use different pre-tokenization patterns.
Both patterns are given below using regex syntax with Unicode property
classes.

\paragraph{Stage 1 (subword).}
Stage~1 uses a conventional subword pre-tokenization pattern. The pattern
separates word-like units, contractions, numeric sequences of one to three
digits, symbols, line breaks, and other whitespace patterns.

\begin{verbatim}
regex = (
    r"[^\r\n\p{L}\p{N}]?[\p{Lu}\p{Lt}\p{Lm}\p{Lo}\p{M}]*[\p{Ll}\p{Lm}\p{Lo}\p{M}]+"
    r"(?i:'s|'t|'re|'ve|'m|'ll|'d)?"
    r"|[^\r\n\p{L}\p{N}]?[\p{Lu}\p{Lt}\p{Lm}\p{Lo}\p{M}]+[\p{Ll}\p{Lm}\p{Lo}\p{M}]*"
    r"(?i:'s|'t|'re|'ve|'m|'ll|'d)?"
    r"|\p{N}{1,3}"
    r"| ?[^\s\p{L}\p{N}]+[\r\n/]*"
    r"|\s*[\r\n]+"
    r"|\s+(?!\S)"
    r"|\s+"
)
\end{verbatim}

\paragraph{Stage 2 (superword).}
The Stage~2 regex augments each word branch of the Stage~1 pattern with a
repeated group of the form \verb|(?: <space><letter-run>)*|. This permits a
pre-tokenization unit to span multiple whitespace-delimited words while
leaving digit grouping, punctuation, line-break, and whitespace handling
unchanged.

\begin{verbatim}
regex = (
    r"[^\r\n\p{L}\p{N}]?[\p{Lu}\p{Lt}\p{Lm}\p{Lo}\p{M}]*[\p{Ll}\p{Lm}\p{Lo}\p{M}]+"
    r"(?: [\p{Lu}\p{Lt}\p{Lm}\p{Lo}\p{M}]*[\p{Ll}\p{Lm}\p{Lo}\p{M}]+)*"
    r"(?i:'s|'t|'re|'ve|'m|'ll|'d)?"
    r"|[^\r\n\p{L}\p{N}]?[\p{Lu}\p{Lt}\p{Lm}\p{Lo}\p{M}]+[\p{Ll}\p{Lm}\p{Lo}\p{M}]*"
    r"(?: [\p{Lu}\p{Lt}\p{Lm}\p{Lo}\p{M}]+[\p{Ll}\p{Lm}\p{Lo}\p{M}]*)*"
    r"(?i:'s|'t|'re|'ve|'m|'ll|'d)?"
    r"|\p{N}{1,3}"
    r"| ?[^\s\p{L}\p{N}]+[\r\n/]*"
    r"|\s*[\r\n]+"
    r"|\s+(?!\S)"
    r"|\s+"
)
\end{verbatim}

\section{Communication-Volume Derivation for Figure~\ref{fig:window-aware-sp}}
\label{app:cp-volume}

This appendix records the per-rank communication-volume expressions used to produce Figure~\ref{fig:window-aware-sp}b and the intensity curves of Figure~\ref{fig:window-aware-sp}c. Volume is not a latency proxy for comparing Ring Attention with Ulysses, for two reasons. First, a balanced Ring rotation can overlap its communication with attention computation, while the Ulysses all-to-all cannot. Second, Ring's volume is not the same on every rank: with early exit, rank $r$ receives $r$ key/value shards, whereas the Ulysses all-to-all is identical on every rank, so the comparison between the two methods reverses direction depending on whether the mean or the busiest rank is taken. The body text therefore does not rest the full-attention algorithm choice on this comparison; the expressions are provided to make the figure reproducible.

The Ring baseline is priced with the causal early exit our implementation performs: a key/value block stops being forwarded once it has passed its last consumer, so rank~$r$ receives only the $r$ shards preceding it and the rank average is $(P-1)/2$. A standard uniform rotation, which forwards every block to every rank, would double these average volumes; the busiest rank is unchanged either way, since the last rank receives every other shard under both. A Ring Attention rotation therefore communicates approximately
$(P-1)L(h_k d_{qk}+h_v d_v)/(2P)$ elements per rank on average, whereas Ulysses communicates
$(P-1)L[(h_q+h_k)d_{qk}+(h_v+h_o)d_v]/P^2$ elements per rank. Their ratio is
\begin{equation}
  \frac{C_{\mathrm{Ring}}}{C_{\mathrm{Ulysses}}}
  = \frac{P(h_k d_{qk}+h_v d_v)}
  {2[(h_q+h_k)d_{qk}+(h_v+h_o)d_v]}.
\end{equation}
Since $h_o=h_q$ by construction and $h_v=h_k$ in our configuration, the ratio simplifies exactly to $P/[2(1+h_q/h_k)]=2/3$: averaged over ranks, Ring moves $33\%$ less data than Ulysses at full attention. The last rank, however, receives every other shard under both a complete rotation and early exit, so for that rank the ratio is instead $P/(1+h_q/h_k)=4/3$ and the ordering reverses. Figure~\ref{fig:window-aware-sp}b shows both: the solid bar is the rank average, and its faded continuation reaches the volume received by the last rank. These expressions count forward-pass communication; the backward pass exchanges the corresponding key and value gradients for Ring Attention and applies the inverse all-to-all operations for Ulysses, adding the same relative volume to both methods and preserving the ratio. The unequal dimensions $d_{qk}=192$ and $d_v=128$ cancel exactly under these head-count equalities. For sliding-window attention, the halo communicates $(P-1)W(h_k d_{qk}+h_v d_v)/P$ elements per rank, so the full-attention-to-halo ratio is $L/(2W)$, which equals $1{,}024$ at $L=256$K and $W=128$.

\section{Evaluation details}
\label{sec:evaluation-details}

\paragraph{$\tau^2$-Bench Telecom}
We use Qwen3-235B-A22B-2507 in non-reasoning mode as the user simulator. Each task repeat is limited to a maximum of 100 steps.

\paragraph{Terminal-Bench 2.1}
We use the Terminus 2 agent harness and limit each run to a maximum of 250 episodes. Each episode corresponds to one cycle in which the model observes the current state and plans a sequence of terminal actions. The per-task timeout is set to two hours, or to the task-specific timeout if it is longer.

\paragraph{SWE-bench Verified}
We use mini-SWE-agent, allowing up to 16K generated tokens per step and limiting each task to a maximum of 250 steps, with a per-task timeout of four hours.

\section*{Contributions}

All authors are alphabetically sorted by last name.

\paragraph{Technical and management leadership.} Joon Son Chung, Sungmin Lee, Junghwan Lim

\paragraph{Core contributors.} Wai Ting Cheung, Gihun Cho, Minsu Ha, Sangho Kang, Beomgyu Kim, Dongseok Kim, Jangwoong Kim, Taehyun Kim, Taewhan Kim, Jeesoo Lee, Jeongdoo Lee, Junhyeok Lee, Dongpin Oh

\paragraph{Contributors.} Hyeyeon Cho, Dahye Choi, Jaeheui Her, Hanbin Jung, Changjin Kang, Minjae Kim, Youngrok Kim, Hyukjin Kweon, Hongjoo Lee, Yeongjae Park, Bokki Ryu

\end{document}